\documentclass[10pt,twocolumn,letterpaper]{article}

\usepackage[pagenumbers]{cvpr} 

\usepackage{url}            
\usepackage{booktabs}       
\usepackage{amsfonts}       
\usepackage{nicefrac}       
\usepackage{microtype}      
\usepackage{xcolor}         
\usepackage{graphicx}
\usepackage{soul,color}
\usepackage{multirow}
\usepackage{rotating}
\usepackage{vcell}
\usepackage{hhline}
\usepackage[shortlabels]{enumitem}
\usepackage{listings}
\usepackage{algorithm,algpseudocode}
\usepackage{amsmath,amsthm,amssymb}
\usepackage{colortbl}
\usepackage{float}
\usepackage{wrapfig}
\usepackage{natbib}
\usepackage{pifont}

\definecolor{mercury}{rgb}{0.898,0.898,0.898}

\newcommand{\cmark}{\textcolor{green}{\ding{52}}}%
\newcommand{\xmark}{\textcolor{red}{\ding{55}}}%

\definecolor{cvprblue}{rgb}{0.21,0.49,0.74}
\usepackage[pagebackref,breaklinks,colorlinks,allcolors=cvprblue]{hyperref}

\title{Enhancing Parameter-Efficient Fine-Tuning of Vision Transformers through Frequency-Based Adaptation}

\author{Son Thai Ly and Hien V. Nguyen\\
University of Houston\\
{\tt\small \{stly,hvnguy35\}@cougarnet.uh.edu}
}

\begin{document}
\maketitle


\begin{abstract}
Adapting vision transformer foundation models through parameter-efficient fine-tuning (PEFT) methods has become increasingly popular. These methods optimize a limited subset of parameters, enabling efficient adaptation without the need to fine-tune the entire model while still achieving competitive performance. 
However, traditional PEFT methods may limit the model's capacity to capture complex patterns, especially those associated with high-frequency spectra. This limitation becomes particularly problematic as existing research indicates that high-frequency features are crucial for distinguishing subtle image structures.
To address this issue, we introduce FreqFit, a novel \underline{Freq}uency \underline{Fi}ne-\underline{t}uning module between ViT blocks to enhance model adaptability. FreqFit is simple yet surprisingly effective, and can be integrated with all existing PEFT methods to boost their performance. By manipulating features in the frequency domain, our approach allows models to capture subtle patterns more effectively. Extensive experiments on 24 datasets, using both supervised and self-supervised foundational models with various state-of-the-art PEFT methods, reveal that FreqFit consistently improves performance over the original PEFT methods with performance gains ranging from 1\% to 16\%. For instance, FreqFit-LoRA surpasses the performances of state-of-the-art baselines on CIFAR 100 by more than 10\% even without applying regularization or strong augmentation. For reproducibility purposes, the source code is available at \href{https://github.com/tsly123/FreqFiT}{https://github.com/tsly123/FreqFiT}.
\end{abstract}

\begin{figure}[t]
\centering
  \begin{minipage}[b]{0.85\columnwidth}
  \centering
    \includegraphics[width=\columnwidth]{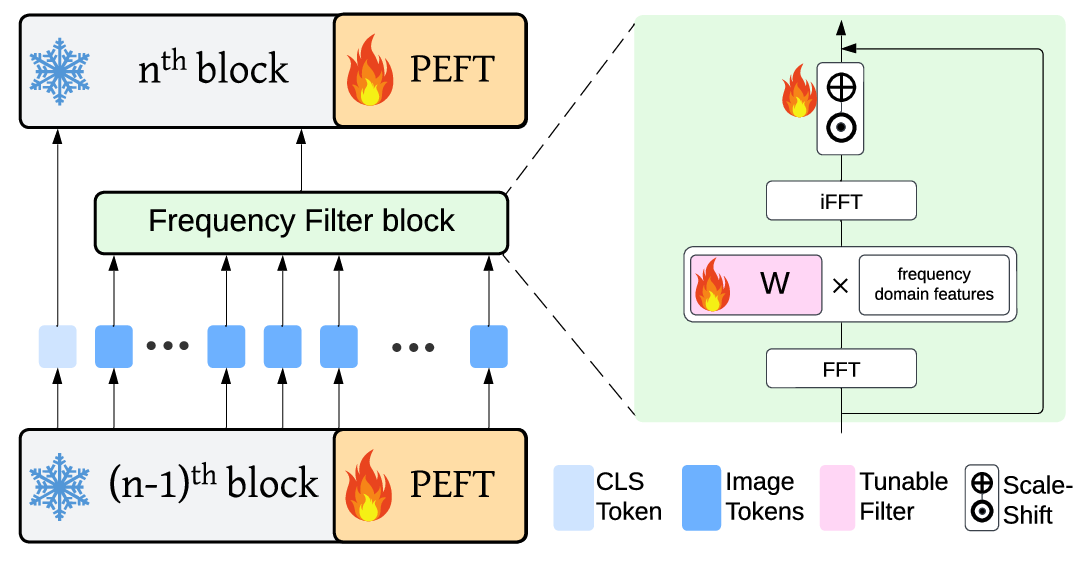} 
    \subcaption{Overview of FreqFit}
  \end{minipage}
  \\
  \begin{minipage}[b]{\columnwidth}
  \centering
    \includegraphics[trim={0 0 0 20},clip,width=0.48\linewidth]{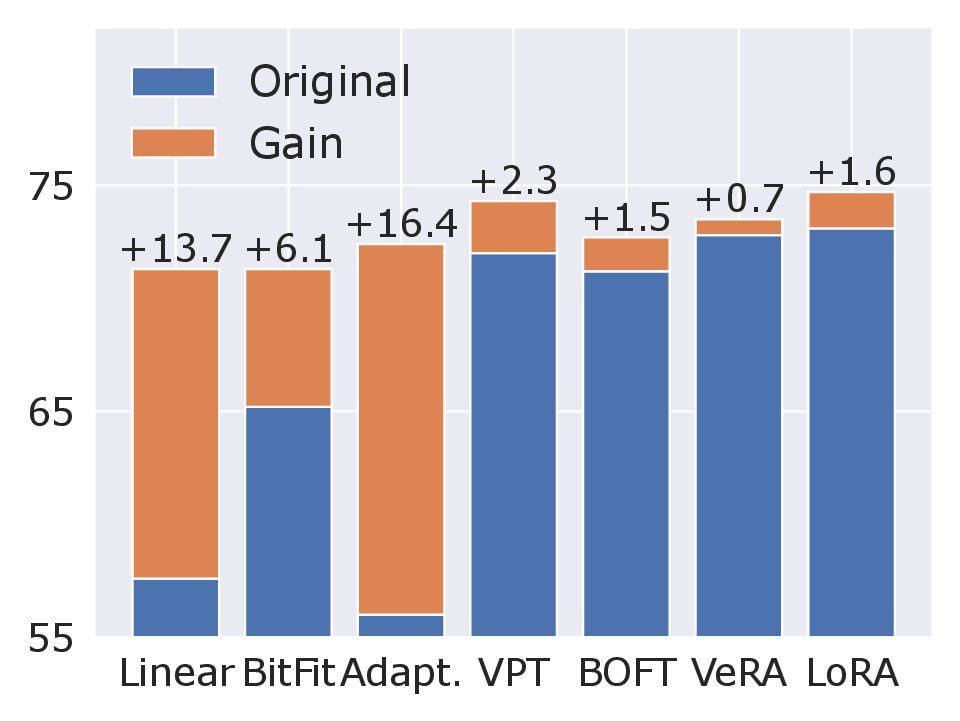} 
    \includegraphics[trim={0 0 0 10},clip,width=0.48\linewidth]{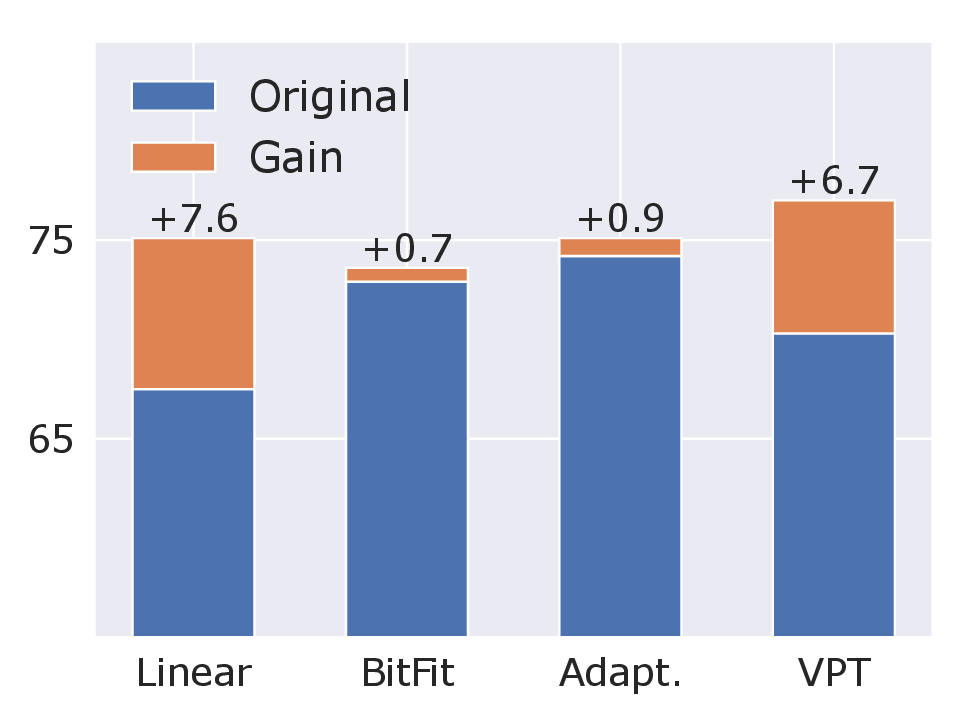} 
    \subcaption{Performance gains with (left) Imagenet-21K and (right) MoCo}
\end{minipage}
\caption{(a) Overview of FreqFit.
(b) Performance gains while incorporating FreqFit of various PEFT methods with pre-trained supervised Imagenet-21K and self-supervised MoCo on VTab-1K.
 } \label{fig:fig1}
\end{figure}

\section{Introduction}
The availability of transformer-based \textit{foundation models} has revolutionized the domain adaptation research. Many fine-tuning methods have been proposed to effectively leverage the good representation of foundation models \citep{vpt,bias1,bias2,scaleshift,continual,adaptformer,lora}. These parameter-efficient fine-tuning (PEFT) methods work by focusing on a small subset of parameters. This approach retains the original pre-trained parameters for the most part, tuning only targeted operations, which dramatically reduces computational overhead while maintaining competitive performance on downstream tasks. For example, BitFit \cite{bias1,bias2} updates only the bias term of the pre-trained backbone, Adapter \cite{adapter1} and AdaptFormer \cite{adaptformer} insert bottleneck-like MLP modules with residual connection inside ViT's blocks, or Lora \cite{lora} injects trainable rank decomposition matrices into each layer of the Transformer architecture.

However, many PEFT methods may limit the model's capacity to capture complex patterns, especially those associated with high-frequency spectra. The limitation becomes particularly problematic as existing research indicates that high-frequency features are vital for improving performance \cite{fft1,rao2021global,inceptrans,rankfourier,vitprop6,vitwork}. These high-frequency components play a key role in tasks that require a deep understanding of intricate image details, such as fine-grained classification, object detection, and medical imaging. Without effective modeling of these high-frequency patterns, PEFT methods risk underperforming, particularly when applied to complex, real-world datasets where such subtle distinctions are essential for accurate predictions.

While it might be argued that PEFT methods have indirectly addressed this frequency limitation by tuning the necessary operations, including self-attentions and other linear layers, our empirical and theoretical findings suggest otherwise. 
As illustrated in Fig. \ref{fig:fig1}, incorporating the feature transformation modules between ViT blocks improves the performance over the original PEFT method.

Building on these insights, we propose FreqFit, a straightforward frequency fine-tuning method designed to modify the features' spectra before it passes through the subsequent ViT blocks. As shown in Fig. \ref{fig:fig1}, FreqFit begins by performing a Fast Fourier Transform (FFT) along the spatial dimensions to convert features into the frequency domain. The spectra are then modulated using a learnable filter. After modulation, the spectral features are converted back to the spatial domain with an inverse FFT (iFFT), followed by a learnable scaling and shifting module, and finally added as the residual connection to the original input. 
 
Our main contributions are as follows:
\begin{enumerate}
\item We propose FreqFit, a simple and effective frequency-based fine-tuning module that seamlessly integrates with existing PEFT methods to enhance model adaptation.
\item We provide theoretical support for why FreqFit can capture image features that existing PEFT methods cannot.
\item We provide a detailed analysis using 24 diverse datasets to demonstrate that FreqFit’s frequency modulation improves ViT token representations and model adaptability.
\end{enumerate}

\section{Related Works}\label{sec:related}

\textbf{Parameter-efficient fine-tuning Methods} Fine-tuning knowledge from pre-trained or foundation models has emerged as a quick and efficient approach to learning new tasks. Recent studies on fine-tuning methods could be categorized into two main approaches: (i) Adopting the subset of tunable input to frozen pre-trained models with Visual Prompt-tuning \citep{vpt} as the representative method. Visual Prompt-tuning (VPT) emerges as a promising solution to address the challenges of domain adaptation. By tuning a small set of additional input tokens, VPT alternates the input domain to optimally align with the frozen pre-trained model. 
(ii) Minimally tuning a small subset of parameters of the pre-trained model while keeping the rest unaltered. The noticeable methods can be mentioned as Bias \citep{bias1,bias2} simply fine-tune solely the Bias terms of pre-trained models \citep{bias1,bias2,scaleshift}, Adapter \citep{adapter1,adapter2,adapter3} inserts bottleneck-like modules with residual connections into ViTs backbone, and low-rank method as LoRA \cite{lora}, BOFT \cite{boft}, VeRA \cite{vera}, FourierFT \cite{fourierft}.
Both VPT and minimal weight-tuning have demonstrated their efficacy and been explored in many directions, such as long-tailed image classification \citep{longtail}, adversarial attacks \citep{adversarial1,adversarial2}, generative visual \citep{generative1,generative2}, point cloud analysis \citep{pointcloud,pointcloud2}, and continual learning \citep{continual,continual2,video}.

However, these methods greatly rely on hyper-parameters such as the number of prompts and the reduction factor in the case of VPT and Adapter, respectively. Their performances show inconsistency across various hyper-parameters settings \citep{vpt,mvlpt,continual,adaptformer,adapterdense,promptsearch,adapterconv}. For example, the effectiveness of VPT significantly depends on the prompt tokens hyper-parameters as they directly determine how prompt tokens interact with image tokens in the spatial domain \citep{vpt,continual,continual2,vptssl}. Another problem is that in the fine-tuning context, the pre-trained backbone parameters are frozen, including self-attention which data-dependably captures long-term dependencies. The existing fine-tuning methods which mainly perform on spatial domain have not been able to learn the equivalent knowledge as self-attention does. Our method takes another approach by modifying the spectrum of the input features to adapt to the frozen pre-trained model in the frequency domain. The learnable filter in our FreqFiT can cover all frequency signals; therefore, it can capture both long-term and short-term interaction between tokens. A frequency-based PEFT method is FourierFT \cite{fourierft} which aims to further compress trainable parameters compared to LoRa \cite{lora} by enjoying the powerful expressiveness of the Fourier transform. Our paper differentiates itself from FourierFT by focusing on the ability to modulate the frequency signal.

\textbf{Fourier Transform in Vision} Fourier transform has been an important tool in digital image processing. There are a variety of works that incorporate the Fourier transform into the deep learning method which suggested the connection between frequency information and superior performances \citep{fft1,rao2021global,fft2,fft3,fft4,fft5,fft6,fft7}. Some of these works utilize the convolution theorem to accelerate the CNNs via FFT computational advantages \citep{fft5,fft6,fft7}. With the advancements of ViTs, there are lines of works that employ the Fourier transform to develop self-attention alternatives \citep{fft3,rao2021global,fft2,fnet,fnet2}. However, these prior works mainly apply in the pre-training state where the backbone's parameters are fully updated, including self-attention.

\section{Methodology}
\textbf{Background on Fourier Transform.} This section provides the background knowledge of Fourier Transform (FT) to set the foundation for understanding the proposed methods. FT has been widely used to decompose signals, such as images and audios, into their consituents frequency components and amplitudes. Concretely, FT for continuous time-domain signal is defined as follows:
\begin{equation}
F(\omega) = \int_{-\infty}^{\infty} f(t) e^{- i \omega t} dt
\end{equation}
where $f(t)$ is the time-domain signal, $F(\omega)$ is the frequency-domain representation of the signal, $\omega$ is the angular frequency, $e^{-i\omega t} = \cos{(\omega t)} - i \sin{(\omega t)}$ is a complex exponential function. The output of this transformation $F(\omega)$ is a complex function encoding both amplitude and phase of each frequency in the original signal. One can also convert the frequency-domain representation back to the orignal time-domain signal using inverse FT. 
\begin{equation}
f(t) = \frac{1}{2 \pi} \int_{-\infty}^{\infty} F(\omega) e^{i \omega t} d\omega
\end{equation}
The original FT is typically defined for time-domain signals, but it can also be applied to image features by replacing the time index with the spatial index.

\textbf{Fast Fourier Transform (FFT)}.
In many practical applications, signals are represented as discrete data points rather than continuous functions. Discrete Fourier Transform (DFT) is the version of the FT that is applied to discrete signals, such as time-series data or digital images. The DFT is defined for a sequence of $n$ discrete values, $f[n]$, where $n=0, 1, \ldots, N-1$ as follows:
\begin{align}
F[k] &= \sum_{n=0}^{N-1} f[n] e^{-\omega k n}, \quad
f[n] &= \sum_{k=0}^{N-1} F[k] e^{\omega k n} \label{eq:dft}
\end{align} 
where $\omega=2 i \pi/N$. DFT can be computationally expensive, particularly for large signals, since its direct computation involves $\mathcal{O}(N^2)$ operations. FFT exploits the symmetrical property within DFT to improve the computational efficiency to $\mathcal{O}(N \log{N})$. Otherwise, FFT computes the same features as DFT. Note that Eq.~\ref{eq:dft} can be written in matrix form, which will be used in our theoretical proof of Theorems 1 \& 2 in subsequent sections. 

\textbf{FreqFit - Frequency Fine-tuning.}
Here, we formally introduce the frequency tuning method, called FreqFit, as illustrated in Fig. \ref{fig:fig1}. FreqFit integrates a frequency operator $H(\cdot)$, consisting of a filter basis followed by a residual connection, between ViT blocks. Given an input token $X \in \mathbb{R}^{H \times W \times D}$, we perform FFT along the spatial dimensions to transform the input into $X_{c}$, as shown in Eq.~\ref{eq:fft}. $X_{c}$ is a complex tensor representing the spectrum of $X$ in the frequency domain. We modulate $X_{c}$ by multiplying it with the learnable filter $K \in \mathbb{R}^{H \times W \times D}$, which has the same dimensions as $X_c$, as shown in Eq.~\ref{eq:mul}. Finally, we convert the modulated spectral features $\tilde{X}$ back to the spatial domain using the inverse FFT, as shown in Eq.~\ref{eq:ifft}, and add a residual connection from the original input $X$. This process can be mathematically summarized as follows:
{\small
\begin{align}
    X_c &= \mathcal{F}(X) \in \mathbb{C}^{H \times W \times D} \label{eq:fft} \\
    \widetilde{X_c} &= K \odot X_c  \label{eq:mul}\\
    \widetilde{X} &\leftarrow \mathcal{F}^{-1}(\widetilde{X_c}) \label{eq:ifft} \\
    \widetilde{X} &= \alpha \odot \widetilde{X} + \beta\\
    \hat{X} &= \widetilde{X} + X \label{eq:residual}
\end{align}
}%
where, $\mathcal{F}$ and $\mathcal{F}^{-1}$ are the fast Fourier transform (FFT) and its inverse. $K \in \mathbb{R}^{H \times W \times D}$ denotes a learnable filter, $\alpha$ and $\beta$ are the scale and shift factors, and $\odot$ is the Hadamard product. 
To facilitate the straightforward incorporation of FreqFiT layer into other fine-tuning methods. The 2D DFT can be viewed as performing 1D DFT on the two dimensions alternatively which also satisfied the conjugate symmetry property $X[(P+H\times W) - u] = X \ast [u]$.

\textbf{How Does FreqFit Improve Performance?}\label{subsec:freqfithow}
The basic idea behind frequency-based tuning FreqFit is to learn the interactions among spatial locations in the frequency domain. Many studies have demonstrated that incorporating high-frequency features leads to better performance \citep{vitprop6,vitwork,rankfourier,inceptrans}. We hypothesize that FreqFit amplifies high-frequency components. The spectral modulation capabilities of FreqFit are visualized and discussed in Fig. \ref{fig:freq} and Sec. \ref{sec:ana}, showing that integrating FreqFit increases the high-frequency features. This observation supports our hypothesis.

As shown in Eq. \ref{eq:mul}, the filter \textit{K} modulates the spectrum of \textit{X} through element-wise multiplication. This modulation controls the strength of different frequency components in the output, allowing for the enhancement or suppression of specific frequency ranges within the token signal. In other words, the result of the element-wise multiplication between the filter and tokens is determined by the filter, which is learned through the back-propagation process. In what follows, we provide the theoretical foundation that underscores the importance of incorporating FreqFit into the existing fine-tuning paradigm.

\textbf{Theorem 1.} FreqFit with $O(1)$ parameters can create a feature transformation that spatial-domain parameter-efficient fine-tuning methods cannot replicate.

\textit{Sketch of Proof:} Due to space constraints, the full proof is provided in the appendix. Here, we outline a proof sketch for better understanding. FreqFit performs a 2D FFT for each token dimension by aggregating statistics across all tokens. In other words, FreqFit operates on the  $H\times W$ dimensions of $X$.  This means that changes induced by the frequency filter depend on these aggregated statistics, affecting all tokens simultaneously. In contrast, spatial-domain PEFT methods do not aggregate statistics across all tokens. Instead, they compute aggregated statistics along the $D$ dimension of $X$ within each individual token. For example, LoRa applies changes using a low-rank matrix or other spatial-domain PEFT methods modify a subset of parameters, which affects all tokens, but without relying on aggregated statistics across tokens. Therefore, spatial-domain PEFT methods cannot replicate the feature changes introduced by FreqFit. $\square$ 

Theorem 1 demonstrates that FreqFit is a missing piece within the current parameter-efficient fine-tuning paradigm. By using only  $\mathcal{O}(1)$ parameters, FreqFit can transform features in ways that existing PEFT methods cannot, thereby enhancing the model's ability to capture more complex patterns.

\textbf{Theorem 2.} Combining FreqFit with spatial-domain PEFT methods can create a feature transformation that cannot be achieved by FreqFit or any spatial-domain PEFT method alone.

\textit{Sketch of Proof:} Since FreqFit computes aggregated statistics within the \emph{same token dimensions}, it cannot replicate the effect of spatial-domain PEFT methods, which aggregate statistics across \emph{all token dimensions}. Together with the result of Theorem 1, this suggests that combining these two methods can yield transformations that neither method can achieve on its own. $\square$

Theorem 2 provides a compelling rationale for combining two complementary approaches: FreqFit and traditional PEFT methods like LoRa and Adapters. By leveraging their distinct strengths, this combination enables a significantly more effective fine-tuning strategy. Our experimental results across 24 diverse datasets strongly validate this theory, demonstrating substantial improvements over using either method alone.

\section{Experimental Settings}
\label{subsec:setup}
\textbf{Pre-Trained Foundation Models.} 
In this study, we experiment with different foundation models that were pre-trained on different datasets and learning approaches, including MAE \citep{mae}, MoCo \citep{mocov3}, and ImageNet-21k \citep{imagenet}. All the foundation models have the same ViT-Base backbones \citep{vit}. We also follow the original configurations, such as image size, patch size, etc.
See Supplementary Material for more details on the experimental settings.

\begin{table*}[!t]
\centering
\resizebox{\textwidth}{!}{
\begin{tabular}{cl|c|ccccccc|cccc|ccccccccc} 
\toprule
 &  &  & \multicolumn{7}{c|}{Natural} & \multicolumn{4}{c|}{Specialized} & \multicolumn{8}{c}{Structured} &  \\
\vcell{} & \multicolumn{1}{c|}{\vcell{ViT-B/16}} & \vcell{\begin{tabular}[b]{@{}c@{}}NO\\Mixup/\\Strong\\Aug.\end{tabular}} & \vcell{\begin{sideways}Cifar100\end{sideways}} & \vcell{\begin{sideways}Caltech101\end{sideways}} & \vcell{\begin{sideways}DTD\end{sideways}} & \vcell{\begin{sideways}Flower102\end{sideways}} & \vcell{\begin{sideways}Pets\end{sideways}} & \vcell{\begin{sideways}SVHN\end{sideways}} & \vcell{\begin{sideways}Sun397\end{sideways}} & \vcell{\begin{sideways}Camelyon\end{sideways}} & \vcell{\begin{sideways}EuroSAT\end{sideways}} & \vcell{\begin{sideways}Resisc45\end{sideways}} & \vcell{\begin{sideways}Retinopathy\end{sideways}} & \vcell{\begin{sideways}Clevr-Count\end{sideways}} & \vcell{\begin{sideways}Clevr-Dist\end{sideways}} & \vcell{\begin{sideways}DMLab\end{sideways}} & \vcell{\begin{sideways}KITTI-Dist\end{sideways}} & \vcell{\begin{sideways}dSpr-Loc\end{sideways}} & \vcell{\begin{sideways}dSpr-Ori\end{sideways}} & \vcell{\begin{sideways}sNORB-Azim\end{sideways}} & \vcell{\begin{sideways}sNORB-Elev\end{sideways}} & \vcell{\begin{sideways}Mean\end{sideways}} \\[-\rowheight]
\printcellmiddle & \multicolumn{1}{c|}{\printcellmiddle} & \printcellmiddle & \printcellbottom & \printcellbottom & \printcellbottom & \printcellbottom & \printcellbottom & \printcellbottom & \printcellbottom & \printcellbottom & \printcellbottom & \printcellbottom & \printcellbottom & \printcellbottom & \printcellbottom & \printcellbottom & \printcellbottom & \printcellbottom & \printcellbottom & \printcellbottom & \printcellbottom & \printcellbottom \\ 
\hline
 & Full & - & 68.9 & 87.7 & 64.3 & 97.2 & 86.9 & 87.4 & 38.8 & 79.7 & 95.7 & 84.2 & 73.9 & 56.3 & 58.6 & 41.7 & 65.5 & 57.5 & 46.7 & 25.7 & 29.1 & 68.9 \\ 
\hline
 & Linear & \cmark & 64.4 & 85.0 & 63.2 & 97.0 & 86.3 & 36.6 & 51.1 & 78.5 & 87.5 & 68.5 & 74.0 & 34.3 & 30.6 & 33.2 & 55.4 & 12.5 & 20.0 & 9.6 & 19.2 & 57.6 \\
 & Bias & \cmark & 72.8 & 87.0 & 59.2 & 97.5 & 85.3 & 59.9 & 51.4 & 78.7 & 91.6 & 72.9 & 69.8 & 61.5 & 55.6 & 32.4 & 55.9 & 66.6 & 40.0 & 15.7 & 25.1 & 65.2 \\
 & Adapter-64 & \cmark & 74.2 & 85.8 & 62.7 & 97.6 & 87.2 & 36.3 & 50.9 & 76.3 & 87.5 & 73.7 & 70.9 & 42.9 & 39.9 & 30.4 & 54.5 & 31.9 & 25.6 & 13.5 & 21.4 & 56.0 \\
 & VPT & \cmark & 78.8 & 90.8 & 65.8 & 98.0 & 88.3 & 78.1 & 49.6 & 81.8 & 96.1 & 83.4 & 68.4 & 68.5 & 60.0 & 46.5 & 72.8 & 73.6 & 47.9 & 32.9 & 37.8 & 72.0 \\
 & LoRa & \cmark & 82.9 & 90.8 & 66.9 & 98.7 & 89.9 & 83.8 & 55.1 & 84.5 & 95.7 & 83.4 & 74.7 & 72.2 & 58.5 & 47.8 & 75.4 & 75.7 & 47.1 & 23.0 & 28.1 & 73.1 \\
 & BOFT & \cmark & 78.7 & 90.5 & 68.4 & 98.5 & 88.3 & 83.2 & 53.5 & 81.2 & 95.9 & 82.7 & 70.3 & 72.2 & 60.4 & 37.3 & 74.1 & 63.9 & 42.9 & 26.0 & 30.6 & 71.2 \\
 & VeRA & \cmark & 80.7 & 90.7 & 68.9 & 98.5 & 89.5 & 84.9 & 53.5 & 81.3 & 95.2 & 83.2 & 73.4 & 73.6 & 60.3 & 43.6 & 77.2 & 72.7 & 46.2 & 28.5 & 30.3 & 72.8 \\
& FourierFT & \cmark & 79.8 & 90.4 & 68.8 & 98.3 & 89.9 & 85.0 & 53.1 & 82.7 & 95.2  & 82.6 & 74.9 & 73.5 & 60.1 & 43.0  & 76.9 & 71.2 & 47.6 & 28.1  & 29.4  & 72.8 \\
 & SSF & \xmark & 69.0 & 92.6 & 75.1 & 99.4 & 91.8 & 90.2 & 52.9 & 87.4 & 95.9 & 87.4 & 75.5 & 75.9 & 62.3 & 53.3 & 80.6 & 77.3 & 54.9 & 29.5 & 37.9 & 75.7 \\
 & NOAH & \xmark & 69.6 & 92.7 & 70.2 & 99.1 & 90.4 & 86.1 & 53.7 & 84.4 & 95.4 & 83.9 & 75.8 & 82.8 & 68.9 & 49.9 & 81.7 & 81.8 & 48.3 & 32.8 & 44.2 & 75.5 \\
 & AdaptFormer & \xmark & 70.8 & 91.2 & 70.5 & 99.1 & 90.9 & 86.6 & 54.8 & 83.0 & 95.8 & 84.4 & 76.3 & 81.9 & 64.3 & 49.3 & 80.3 & 76.3 & 45.7 & 31.7 & 41.1 & 74.7 \\
 & RepAdaptor & \xmark & 72.4 & 91.6 & 71.0 & 99.2 & 91.4 & 90.7 & 55.1 & 85.3 & 95.9 & 84.6 & 75.9 & 82.3 & 68.0 & 50.4 & 79.9 & 80.4 & 49.2 & 38.6 & 41.0 & 76.1 \\ 
\hline
\multirow{8}{*}{\begin{tabular}[c]{@{}c@{}}FreqFit-\\(ours)\end{tabular}} & Linear & \multirow{8}{*}{\cmark} & \textbf{78.5} & \textbf{89.8} & \textbf{70.1} & \textbf{98.6} & \textbf{88.7} & \textbf{76.0} & \textbf{53.9} & \textbf{79.3} & \textbf{94.8} & \textbf{81.5} & 72.9 & \textbf{72.0} & \textbf{60.1} & \textbf{39.6} & \textbf{70.5} & \textbf{71.2} & \textbf{39.0} & \textbf{33.0} & \textbf{34.5} &  \textbf{71.3} (\textcolor{green}{+13.7})\\
 & Bias &  & \textbf{82.4} & \textbf{90.4} & \textbf{69.0} & \textbf{98.9} & \textbf{89.5} & \textbf{78.7} & \textbf{54.9} & \textbf{79.9} & \textbf{95.9} & \textbf{82.0} & \textbf{70.3} & \textbf{64.9} & \textbf{59.7} & \textbf{35.7} & \textbf{72.4} & \textbf{74.8} & \textbf{41.4} & \textbf{24.7} & \textbf{36.3} & \textbf{71.3}
 (\textcolor{green}{+6.1}) \\
 & Adapter-64 &  & \textbf{81.5} & \textbf{90.4} & \textbf{70.5} & \textbf{98.7} & \textbf{89.3} & \textbf{82.3} & \textbf{54.2} & \textbf{81.7} & \textbf{96.1} & \textbf{82.3} & \textbf{73.4} & \textbf{71.4} & 61.2 & \textbf{37.2} & \textbf{63.6} & \textbf{84.2} & \textbf{40.1} & \textbf{26.4} & \textbf{38.7} & \textbf{72.4} (\textcolor{green}{+16.4})\\
 & VPT &  & \textbf{80.1} & \textbf{92.8} & \textbf{68.4} & \textbf{98.5} & \textbf{89.1} & \textbf{86.2} & \textbf{54.6} & \textbf{84.4} & \textbf{96.1} & \textbf{83.9} & \textbf{73.1} & \textbf{73.8} & 59.2 & 45.6 & \textbf{75.8} & \textbf{75.4} & \textbf{47.9} & \textbf{35.2} & \textbf{44.8} & \textbf{74.3} (\textcolor{green}{+2.4})\\
 & LoRa &  & \textbf{83.7} & 90.2 & \textbf{68.8} & \textbf{98.9} & \textbf{90.5} & \textbf{86.3} & \textbf{55.1} & \textbf{85.0} & \textbf{96.1} & \textbf{83.5} & \textbf{74.7} & \textbf{74.5} & \textbf{62.0} & 45.9 & \textbf{78.2} & \textbf{78.7} & \textbf{51.7} & \textbf{31.7} & \textbf{36.1} & \textbf{74.7} (\textcolor{green}{+1.6}) \\
 & BOFT &  & 78.1 & \textbf{91.0} & 68.0 & \textbf{98.6} & \textbf{89.5} & \textbf{86.0} & \textbf{54.3} & 80.3 & \textbf{96.1} & \textbf{83.4} & \textbf{74.0} & \textbf{73.8} & \textbf{60.8} & \textbf{38.2} & \textbf{78.2} & \textbf{73.7} & \textbf{45.0} & \textbf{30.0} & \textbf{33.0} & \textbf{72.7} (\textcolor{green}{+1.5}) \\
& VeRA &  & \textbf{81.3} & 90.4 & \textbf{71.6} & \textbf{98.9} & \textbf{90.3 }& \textbf{85.7} & \textbf{54.6} & 80.6 & \textbf{95.9} & \textbf{83.7} & \textbf{73.4} & 69.4 & \textbf{60.5} & 42.5 & \textbf{77.6} & \textbf{83.1} & 44.0  & \textbf{32.0} & \textbf{33.6} &  \textbf{73.5} (\textcolor{green}{+0.8})\\
& FourierFT &  & \textbf{83.8} & \textbf{91.7} & \textbf{72.1} & \textbf{98.9} & \textbf{90.5} & \textbf{87.5} & \textbf{54.2} & \textbf{84.0} & \textbf{95.2} & \textbf{83.6} & 73.6 & 73.3 & 59.8 & \textbf{46.0} & 76.1 & \textbf{76.8} & \textbf{50.7} &\textbf{30.4}& \textbf{33.3} &  \textbf{74.2} (\textcolor{green}{+1.4}) \\
\bottomrule
\end{tabular}
}
\caption{Comparison between tuning the data with scale-shift and frequency tuning method and their counterpart state-of-the-arts on VTab dataset with Imagenet-21K pre-trained weights. \textbf{Bold} indicates that the performance is better than the original method without tuning the data. The \cmark\:and \xmark\:indicate the results produced by NOT using Mixup \cite{mixup} and strong augmentation, and vice versa, respectively. (\textcolor{green}{$\pm$}) is the performance gain compared to the original methods without applying feature transformation.}\label{tab:vtab21}
\end{table*}

\textbf{Fine-Tuning PEFT Methods.} We apply the feature transformation techniques in conjunction with the following PEFT methods. We selected these methods as they are representative examples within their respective approach families.
\begin{itemize}[leftmargin=10pt]
    \item LINEAR: only update the last linear layer as classification head.
    \item BIAS \citep{bias1, bias2}: update only the bias term of the pre-trained backbone.
    \item ADAPTER \citep{adapter1,adapter2,adapter3}: insert bottleneck-like MLP modules with residual connection inside ViT's blocks.
    \item VPT \citep{vpt,vptssl}: adding learnable tokens along with image tokens as input of ViT's blocks. We follow the number of prompt tokens reported in VPT \cite{vpt}.
    \item LoRA \cite{lora}: decomposes a large matrix into two smaller low-rank matrices in the desired layers. We apply LoRa for all linear layers.
    \item BOFT \cite{boft}: leverages butterfly-structured orthogonal parameterization to reduce trainable parameters. We apply BOFT for all linear layers.
    \item VeRA \cite{vera}: using a single pair of low-rank matrices shared across all layers and learning small scaling vectors instead, reducing the number of trainable parameters. We apply VeRA for all linear layers.
\end{itemize}

\textbf{Downstream Tasks.} Following \citep{vpt,adaptformer,vptssl,sidetune2,scaleshift}, we evaluate FreqFiT on the VTAB-1k Natural \citep{vtab} tasks that contain natural images and the Fine-Grained Visual Classification (FGVC) tasks, including CUB-200-2011 \citep{cub}, NABirds \citep{nabird}, Oxford Flowers \citep{flower}, Stanford Dogs \citep{dog} and Stanford Cars \citep{car}. We follow the dataset split in \citep{vpt}. 

\textbf{Hypeparameter Configurations and Regulations/Augmentation.} Our primary goal is to illustrate the effectiveness of the feature transformation approach rather than to compete directly with state-of-the-art methods. Therefore, we avoid extensive hyperparameter searches, including Adapter reduction rate, VPT prompt tokens, LoRa rank, instead applying a single default configuration across all experiments for consistency. In addition, to clearly demonstrate the effectiveness of the feature transformation approach, following \cite{bias1,vpt}, we do not utilize Mixup \cite{mixup} or strong image augmentation techniques in this study. Further details are provided in Supplementary Material.

\section{Results}\label{sec:res}

\textbf{Effects of incorporating FreqFit.} Tables \ref{tab:vtab21} and \ref{tab:ssl} summarize the fine-tuning performance on VTAB-1k tasks with natural images, demonstrating that methods enhanced with feature transformation techniques consistently outperform recent fine-tuning approaches. Notably, these improvements hold across a range of advanced PEFT techniques—including VPT \citep{vpt}, Bias Tuning \citep{bias1, bias2}, Adapter \citep{adapter1, adapter2, adapter3}, LoRA \cite{lora}, BOFT \cite{boft}, and VeRA \cite{vera}—and pre-trained foundation models such as MAE \citep{mae}, MoCo \citep{mocov3}, and ImageNet-21k \citep{imagenet}. For instance, when applied to ImageNet-21k, FreqFit achieves a mean accuracy gain of 1.5\% to 16\% across 19 VTAB-1k tasks, outperforming state-of-the-art PEFT methods. Note that, to clearly demonstrate the effectiveness of the feature transformation approach, we do not utilize Mixup \cite{mixup} and strong augmentations. Even without those strong performance booster techniques, feature-transformed methods yield competitive results and often surpass baseline performances with these enhancements across multiple tasks. For instance, FreqFit-LoRA surpasses other state-of-the-art baselines on CIFAR100 more than 10\%. Furthermore, the Linear transformation method, when used with the feature transformation module, exceeds the performance of other state-of-the-art techniques and even outperforms full fine-tuning (\textit{FULL}) by 1-2\% on several tasks. These results highlight the crucial of modifying features in the frequency domain in optimizing ViTs across diverse settings.

\begin{table}[t]
\centering
\resizebox{\columnwidth}{!}{
\begin{tabular}{clcccccccc} 
\toprule
\textbf{Pre-trained} & \multicolumn{1}{c}{\textbf{ViT-B/16}} & \textbf{Cifar100} & \textbf{Caltech101} & \textbf{DTD} & \textbf{Flowers102} & \textbf{Pets} & \textbf{SVHN} & \textbf{Sun395} & \textbf{Mean} \\ 
\hline
\multirow{9}{*}{MAE} & Linear & 8.7 & 41.5 & 20.6 & 19.2 & 11.3 & 22.2 & 8.6 & 18.9 \\
 & Bias & 22.4 & 82.6 & 49.7 & 66.2 & 67.7 & 69.0 & 24.3 & 54.6 \\
 & Adapter & 35.1 & 85.0 & 56.5 & 66.6 & 71.3 & 45.0 & 24.8 & 54.9 \\
 & VPT & 8.2 & 55.2 & 58.0 & 39.3 & 45.2 & 19.4 & 21.8 & 36.0 \\
 & Gated-VPT & 23.2 & 75.0 & 58.1 & 62.4 & 60.0 & 36.0 & 20.0 & 47.6 \\ 
\cline{2-10}
 & FreqFiT-Linear & \textbf{35.8} & \textbf{81.4} & \textbf{59.6} & \textbf{63.3} & \textbf{68.3} & \textbf{58.5} & \textbf{18.7} & \textbf{55.1} (\textcolor{green}{+36.2}) \\
 & FreqFiT-Bias & \textbf{24.7} & 81.8 & \textbf{50.8} & \textbf{66.4} & 67.5 & 63.0 & 22.1 & 53.8 (-0.8) \\
 & FreqFiT-Adapter-64 & 31.3 & 81.2 & \textbf{60.9} & 66.4 & 70.9 & \textbf{66.1} & 23.2 & \textbf{57.1} (\textcolor{green}{+2.2})\\
 & FreqFiT-VPT & \textbf{27.6} & \textbf{85.8} & \textbf{59.5} & \textbf{71.4} & \textbf{75.5} & \textbf{74.7} & \textbf{21.2} & \textbf{59.4} (\textcolor{green}{+23.4})\\ 
\hline
\multirow{9}{*}{MoCo} & Linear & 62.9 & 85.1 & 68.8 & 87.0 & 85.8 & 41.8 & 40.9 & 67.5 \\
 & Bias & 65.5 & 89.2 & 62.9 & 88.9 & 80.5 & 82.7 & 40.5 & 72.9 \\
 & Adapter & 73.0 & 88.2 & 69.3 & 90.7 & 87.4 & 69.9 & 40.9 & 74.2 \\
 & VPT & 70.1 & 88.3 & 65.9 & 88.4 & 85.6 & 57.8 & 35.7 & 70.3 \\
 & Gated-VPT & 73.5 & 90.0 & 70.8 & 89.7 & 87.0 & 75.0 & 41.5 & 74.8 \\ 
\cline{2-10}
 & FreqFiT-Linear & \textbf{67.3} & \textbf{90.4} & 68.6 & \textbf{87.9} & 85.4 & \textbf{83.7} & \textbf{42.1} & \textbf{75.1} (\textcolor{green}{+7.6})\\
 & FreqFiT-Bias & 63.3 & \textbf{91.0} & \textbf{65.0} & 87.0 & \textbf{85.4} & \textbf{82.9} & 40.4 & \textbf{73.6} (\textcolor{green}{+0.7})\\
 & FreqFiT-Adapter-64 & 65.4 & \textbf{90.2} & 68.8 & 88.8 & 86.5 & \textbf{85.2} & \textbf{41.1} & \textbf{75.1} (\textcolor{green}{+7.6})\\
 & FreqFiT-VPT & \textbf{73.9} & \textbf{90.6} & \textbf{70.4} & \textbf{91.2} & \textbf{88.7} & \textbf{82.2} & \textbf{42.4} & \textbf{77.0} (\textcolor{green}{+6.7})\\ 
\hline
\multirow{3}{*}{CLIP} & FreqFiT-Linear & 69.8 & 92.4 & 75.4 & 96.0 & 88.1 & 57.0 & 53.7 & 76.1 \\
 & FreqFiT-Bias & 18.3 & 82.7 & 53.8 & 68.0 & 59.5 & 45.0 & 46.0 & 53.3 \\
 & FreqFiT-VPT & 74.5 & 93.7 & 77.2 & 96.8 & 91.8 & 82.13 & 56.6 & 81.8 \\
\bottomrule
\end{tabular}
}
\caption{Performance comparison on Vtab-1K Natural with pre-trained self-supervised learning MAE and MoCo. We also present the results with pre-trained CLIP \cite{clip}. \textbf{Bold} indicates FreqFit-method surpasses the original method.}\label{tab:ssl}
\end{table}

\textbf{Supervised versus self-supervised pre-trained models.} Tab. \ref{tab:vtab21} and \ref{tab:ssl} show that FreqFit methods yield better performance gains when applied to supervised pre-trained models, such as ImageNet-21K, compared to self-supervised pre-trained models like MAE and MoCo. Specifically, FreqFit techniques consistently produce higher accuracy improvements with MoCo than MAE pre-trained models. The results highlight the importance of addressing the limitation in capturing frequency patterns regardless the pre-training strategy.

Next, we will discuss how incorporating FreqFit improves the performance for the representative PEFT methods.

\noindent \textbf{- FreqFit with VPT.} Our simple yet effective method outperforms the original VPT by 23\% and 7\% on average across all Natural tasks for pre-trained MAE and MoCo, respectively, and 2.4\% on all tasks with Imagenet-21K. We hypothesize that this superiority is due to the frequency properties of FreqFiT. Typically, VPT learns the interaction among tokens in the spatial domain by appending prompt tokens to image tokens, with long-term dependencies learned by the self-attention module \citep{vaswani2017attention,rank,rao2021global}. This important knowledge may not be fully captured when all the backbone's parameters, including self-attention, are not updated for new tasks under VPT framework. In contrast, our FFT-based tuning approach captures both long-term and short-term interactions. It is worth noting our FreqFiT-VPT does not search for the \textit{optimal prompt length} or the ViT \textit{blocks to insert} prompt tokens as in VPT \citep{vpt}. Instead, we reuse the prompt lengths reported in VPT \citep{vpt} and Gated-VPT \citep{vptssl}, inserting the FreqFiT layer before every ViT block.

\noindent \textbf{- FreqFit with Adapter.} FreqFiT significantly enhances the performance of the original Adapter method, achieving average improvements of 16.4\% across all VTab-1k tasks with Imagenet-1K pre-trained weights. This improvement can be attributed to the original Adapter's limitations in learning long-term information. In the Adapter approach, only the bottleneck-like MLP modules are updated, while all other parameters, including the self-attention modules, remain frozen. Furthermore, we observe performance fluctuations when varying the reduction factor. For a detailed analysis of experiments on prompt lengths and reduction factors, please refer to Sec. \ref{sec:ana}.

\noindent \textbf{- FreqFit with Bias Tuning.} Bias tuning \citep{bias1,bias2} is a competitive, parameter-efficient tuning baseline. Interestingly, despite yielding superior results with 6.1\% gain in performance with Imagenet-21k \cite{imagenet} pre-trained model, the results with MAE \cite{mae} and MoCo \cite{mocov3} do not enhance performance. This suggests the distinction between supervised and self-supervised pre-trained models could have different effects on the performances. Unlike VPT and Adapter, bias tuning is the only baseline that modifies self-attention (bias terms), which is essential for learning long-term dependencies.
Since only the bias term is updated, this strategy can be viewed as a linear transformation, which may not be sufficient to handle complex data distribution shifts or effectively capture the frequency patterns. Consequently, we hypothesize that the linearly shifted outputs from bias tuning could negatively impact the long-term dependency-capturing ability of the FreqFiT layer. To better understand this phenomenon, we will present a visualization of the frequency filter and discuss it in the subsequent section.

\begin{figure}[t]
  \begin{minipage}[]{\columnwidth}
    \centering{Imagenet-21K - CIFAR100}\\
    \centering
    \includegraphics[width=0.45\columnwidth]{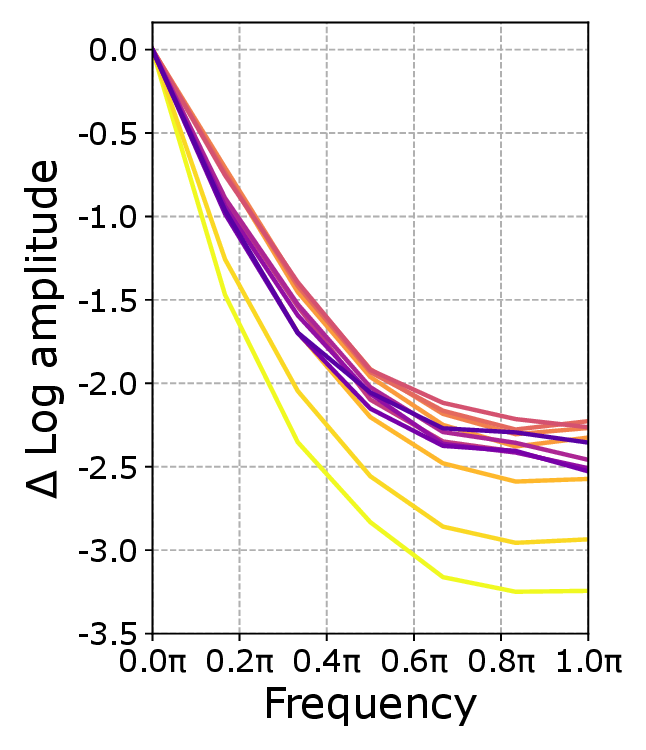} 
    \hspace{-10pt}
    \includegraphics[width=0.45\columnwidth]{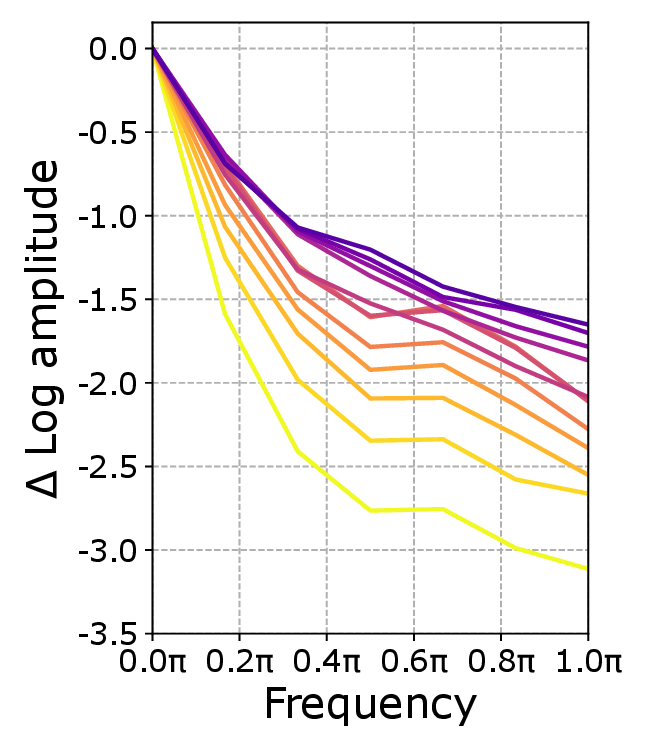}    
    \includegraphics[width=0.1\columnwidth]{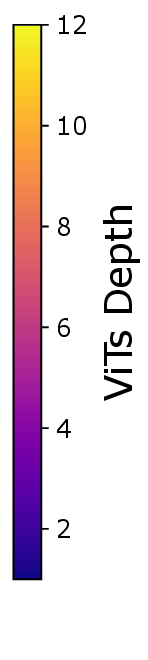}
    \centering{(Left) LoRA. (Right) FreqFiT-LoRA}
  \end{minipage}
\caption{Relative log amplitudes of Fourier transformed feature maps. $\Delta$ Log amplitude means relative logarithmic amplitude concerning the logarithmic amplitude at normalized frequency $0\pi$ (center) and $1\pi$ (boundary). The brighter the color, the deeper the layer. Our FreqFiT has higher amplitudes compared to the original methods. These visualizations suggest our FreqFit is better at capturing high-frequency components, and potentially leading to better performance. More in Supplementary Material.} \label{fig:freq}
\end{figure}

\noindent \textbf{- FreqFiT with low-rank methods.} LoRA \cite{lora} and its variants, BOFT \cite{boft} and VeRA \cite{vera}, reparameterize a large matrix into smaller low-rank matrices. In this study, we use the default hyper-parameter configurations from HuggingFace \cite{huggingface} and apply these PEFT techniques for all linear layers for all experiments. FreqFiT consistently enhances the performance of these methods, achieving average improvements of 1.6\%, 1.5\%, and 0.8\% across all VTab-1k tasks with Imagenet-1K pre-trained weights with LoRA, BOFT, and VeRA, respectively. Since we apply these methods for all linear layers, including those of self-attention operations, they may be able to capture the spatial dependencies for new tasks and address the inter-block relationship problems. However, when incorporated with FreqFit, these PEFT methods achieve better performances, validating our approach of altering the feature in the frequency domain.

\begin{figure}[t]
\begin{minipage}[]{0.48\linewidth}
  \centering{\footnotesize{MAE/FreqFiT-Adapter/Caltech101}}
    \includegraphics[trim=55 0 40 15,clip,width=\textwidth]{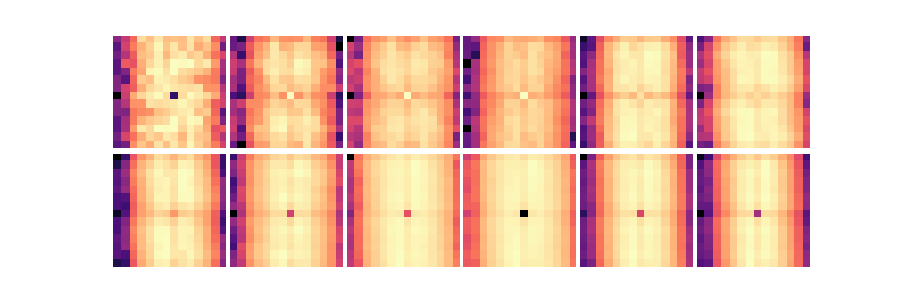} 
\end{minipage}
\hfill
\begin{minipage}[]{0.48\linewidth}
  \centering{\footnotesize{MAE/FreqFiT-Bias/Caltech101}}
    \includegraphics[trim=55 0 40 15,clip,width=\textwidth]{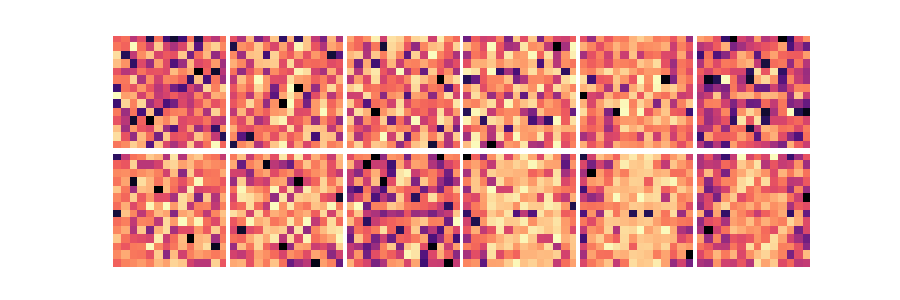} 
\end{minipage}\\
\vskip -3pt
\begin{minipage}[]{0.48\linewidth}
  \centering{\footnotesize{{CLIP/FreqFiT-VPT/Flower102}}}
    \includegraphics[trim=55 0 40 15,clip,width=\textwidth]{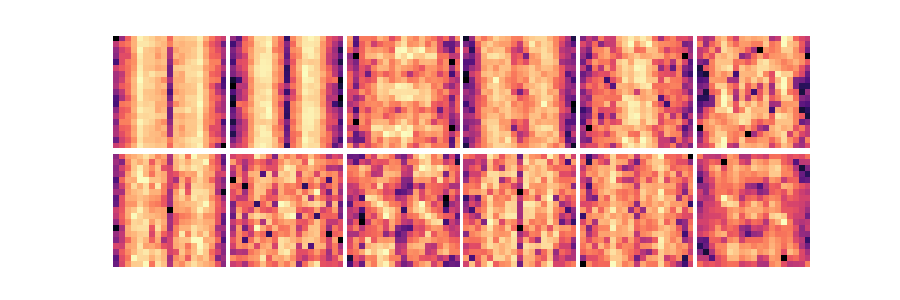} 
\end{minipage}
\hfill
\begin{minipage}[]{0.48\linewidth}
    \centering{\footnotesize{{CLIP/FreqFiT-Bias/Flower102}}}
    \includegraphics[trim=55 0 40 15,clip,width=\textwidth]{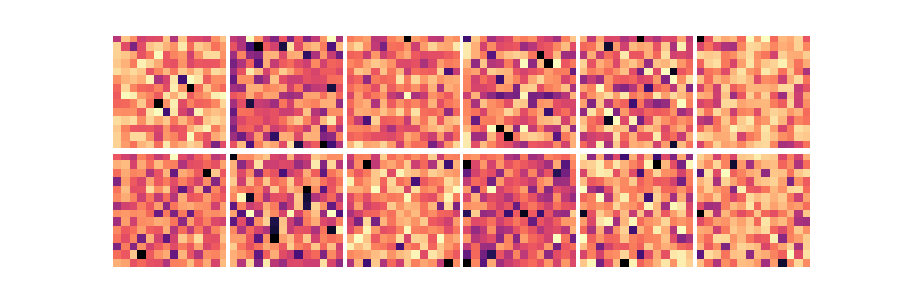}    
\end{minipage}\\

\vskip -3pt
\begin{minipage}[]{0.48\linewidth}
  \centering{\footnotesize{{Imagenet/FreqFiT-Lora/Cifar100}}}
    \includegraphics[width=\textwidth]{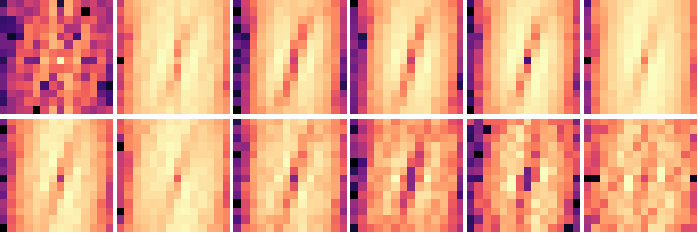} 
\end{minipage}
\hfill
\begin{minipage}[]{0.48\linewidth}
    \centering{\footnotesize{{Imagenet/FreqFiT-Bias/Cifar100}}}
    \includegraphics[width=\textwidth]{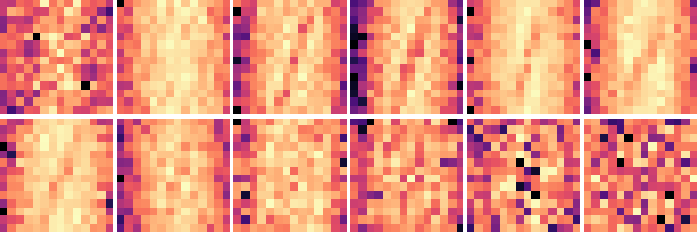}    
\end{minipage}
\caption{Filters in the frequency domain in different fine-tuning settings. For each case, we show 12 filters that are inserted before the $1^{th}$ to $12^{th}$ blocks of ViTs, which are labeled according to ``\textit{pre-trained model/incorporated method/task}''. The center pixel represents the zero frequency and the brighter pixel indicates higher amplitudes. Our visualization shows that the incorporated FreqFiT-VPT, FreqFiT-Adapter, and FreqFit-LoRA can capture high-frequency components by adopting our FreqFiT. Whereas, the FreqFiT-Bias does not show a clear pattern of capturing high-frequency components. 
} \label{fig:filter}
\end{figure}

\section{Ablation Study}\label{sec:ana}

\textbf{Frequency analysis.} Given that FreqFit operates primarily in the frequency domain, it is essential to analyze the learned transformations. Following \citep{vitwork,fft2,inceptrans}, we compare the relative log amplitudes of the Fourier transform of the output feature maps. Fig. \ref{fig:freq} visualizes the differences in the relative log amplitudes of Fourier-transformed feature maps between (left) original LoRA and (right) FreqFit-LoRA, the FreqFiT-enhanced versions. The $\Delta$ log amplitude represents the relative logarithmic amplitude between frequency $0\pi$ (center) and $1\pi$ (boundary). Brighter colors indicate deeper layers. 

Across various tasks and fine-tuning methods, a common pattern emerges: FreqFiT tends to increase the amplitude, as hypothesized in Sec. \ref{subsec:freqfithow}. Since high-frequency features are important in capturing interactions among spatial locations in the frequency domain, leading to better performance as discussed in \citep{fft1,rao2021global,inceptrans,rankfourier,vitprop6,vitwork}, it is logical that our method increases the amplitude. FreqFiT filters can address both low and high frequencies, enabling the FreqFiT-enhanced methods to capture high-frequency features more effectively than the original methods, as evidenced by the higher amplitudes shown in Fig. \ref{fig:freq}.



%

To better understand this behavior, we visualize the filters of different fine-tuning settings, as shown in Fig. \ref{fig:filter}. Our visualization shows that the incorporated FreqFiT-LoRA, FreqFiT-VPT, and FreqFiT-Adapter can capture high-frequency components by adopting our FreqFiT. In addition, we can also see high-pass, low-pass, and band-pass filters in the visualization. This is reasonable as the filters in our FreqFiT cover all frequency ranges. Regarding the FreqFiT-Bias visualizations, they do not show a clear pattern of capturing high-frequency components. This visualization justifies our hypothesis and results shown in Sec. \ref{sec:res}. 

Compared to VPT and Adapter, Bias tuning is the only baseline that modifies the self-attention mechanism. However, its effectiveness is lower than the other methods. This suggests that Bias tuning and our FreqFiT are not necessarily complementary. Importantly, our FreqFiT captures both low and high frequencies. Previous studies \citep{fft1,fft2,vitwork,vitprop6} have indicated a link between high-frequency components in tokens and improved performance. These findings validate our frequency-tuning approach and highlight a potential research direction: adaptive frequency-tuning, which could result in filters functioning as high-pass filters. On the other hand, we can recognize a pattern in the learned filter of Bias with the pre-trained Imagenet-1K. However, it is also less obvious compared to that of LoRA. This could be because the difference in pre-trained foundation model, suggesting the drawback of linear transformation in capturing useful frequency signals.

\begin{figure}[t]
\begin{minipage}[]{\columnwidth}
\centering
    \includegraphics[width=0.7\columnwidth]{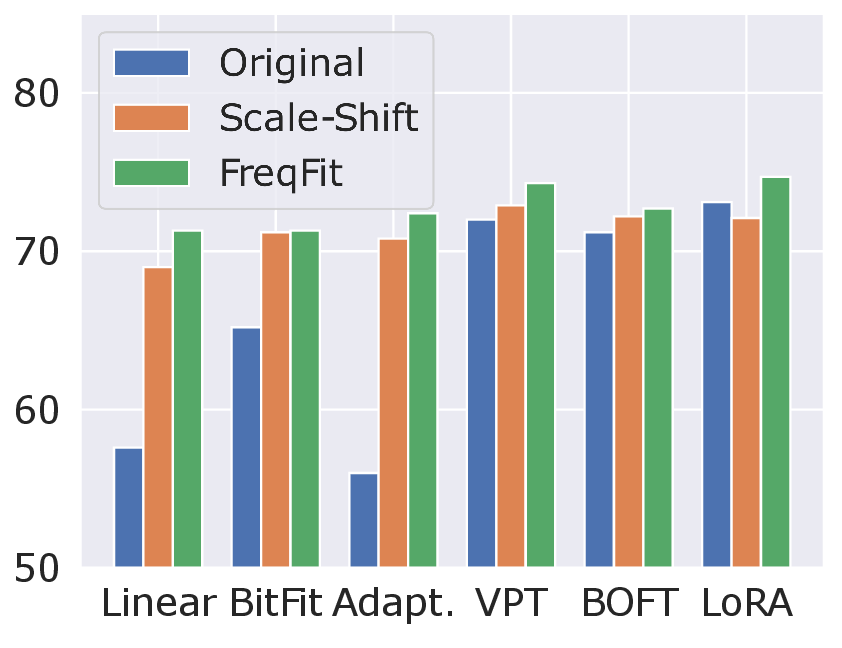}
\end{minipage}
\caption{Performances when applied FreqFit outperform when applied Scale-Shift on 19 tasks of VTab-1K with Imagenet-21K pre-trained model.}\label{fig:ssf}
\end{figure}

\textbf{FreqFit versus Scaling-Shifting.} Fig. \ref{fig:ssf} presents a detailed comparison of two feature transformation techniques: Scaling-Shifting and frequency tuning (FreqFit). The results indicate that FreqFit consistently outperforms Scaling-Shifting, with a mean performance gain difference of 1.2\% across all evaluated methods, underscoring the effectiveness of FreqFit in aligning input features more closely with the frozen model parameters. This superiority may be attributed to FreqFit’s ability to operate across the all frequencies spectrum, enabling it to capture a wider range of spatial dependencies within the token representations. By modulating frequencies, FreqFit can more effectively capture both long-term structural patterns and short-term fine-grained details, providing a more holistic transformation that allows the frozen parameters to better adapt to variations in data distribution. Furthermore, prior research has underscored a strong connection between high-frequency signal capture and improved model performance \cite{rao2021global,vitwork,fft1,fft2,fft3,fft4,fft5,fft6,fft7}. FreqFit’s frequency modulation allows it to harness these high-frequency signals effectively, thus reinforcing its potential to capture subtle spatial relationships and complex data patterns. This positions FreqFit as a promising enhancement for parameter-efficient fine-tuning, offering frozen models a level of flexibility and robustness in adapting to new tasks that Scaling-Shifting alone cannot provide.

\begin{table}[t]
\centering
\resizebox{\columnwidth}{!}{
\setlength{\extrarowheight}{-1pt}
\addtolength{\extrarowheight}{\aboverulesep}
\addtolength{\extrarowheight}{\belowrulesep}
\setlength{\aboverulesep}{0pt}
\setlength{\belowrulesep}{0pt}
\begin{tabular}{clccccccc} 
\toprule
 & \multicolumn{1}{c}{\textbf{ViT-B/16}} & \multicolumn{1}{l}{\textbf{\#runs}} & \textbf{CUB} & \textbf{Flower} & \textbf{NABird} & \textbf{Dog} & \textbf{Car} & \textbf{Mean} \\ 
\hline
\multirow{5}{*}{MAE} & {\cellcolor[rgb]{0.898,0.898,0.898}}VPT & {\cellcolor[rgb]{0.898,0.898,0.898}}- & {\cellcolor[rgb]{0.898,0.898,0.898}}68.3 & {\cellcolor[rgb]{0.898,0.898,0.898}}80.1 & {\cellcolor[rgb]{0.898,0.898,0.898}}65.2 & {\cellcolor[rgb]{0.898,0.898,0.898}}78.8 & {\cellcolor[rgb]{0.898,0.898,0.898}}67.7 & {\cellcolor[rgb]{0.898,0.898,0.898}}72.0 \\
 & {\cellcolor[rgb]{0.898,0.898,0.898}}Gated-VPT & {\cellcolor[rgb]{0.898,0.898,0.898}}- & {\cellcolor[rgb]{0.898,0.898,0.898}}70.6 & {\cellcolor[rgb]{0.898,0.898,0.898}}78.6 & {\cellcolor[rgb]{0.898,0.898,0.898}}67.3 & {\cellcolor[rgb]{0.898,0.898,0.898}}78.9 & {\cellcolor[rgb]{0.898,0.898,0.898}}71.7 & {\cellcolor[rgb]{0.898,0.898,0.898}}73.4 \\
 & Gated-VPT w/ seeds & 5 & 68.8 (1.2) & 76.1 (1.0) & 65.5 (2.4) & 67.6 (15) & 71.8 (2.0) & 70.0 (4.3) \\ 
\cline{2-9}
 & \multirow{2}{*}{FreqFiT-VPT (ours)} & 3 & 74.7 & 83.3 & 69.3 & 79.2 & 74.5 & 76.2 \\
 &  & 5 & 74.2 (0.9)~ & 82.5 (1.4) & 68.9 (0.6) & 78.8 (0.4) & 73.4 (1.9) & 75.5 (1.0) \\ 
\hline
\multirow{5}{*}{MoCo} & {\cellcolor[rgb]{0.898,0.898,0.898}}VPT & {\cellcolor[rgb]{0.898,0.898,0.898}}- & {\cellcolor[rgb]{0.898,0.898,0.898}}82.7 & {\cellcolor[rgb]{0.898,0.898,0.898}}94.4 & {\cellcolor[rgb]{0.898,0.898,0.898}}76.0 & {\cellcolor[rgb]{0.898,0.898,0.898}}83.3 & {\cellcolor[rgb]{0.898,0.898,0.898}}79.2 & {\cellcolor[rgb]{0.898,0.898,0.898}}83.1 \\
 & {\cellcolor[rgb]{0.898,0.898,0.898}}Gated-VPT & {\cellcolor[rgb]{0.898,0.898,0.898}}- & {\cellcolor[rgb]{0.898,0.898,0.898}}82.9 & {\cellcolor[rgb]{0.898,0.898,0.898}}93.7 & {\cellcolor[rgb]{0.898,0.898,0.898}}76.0 & {\cellcolor[rgb]{0.898,0.898,0.898}}83.4 & {\cellcolor[rgb]{0.898,0.898,0.898}}79.0 & {\cellcolor[rgb]{0.898,0.898,0.898}}83.0 \\
 & Gated-VPT w/ seeds & 5 & 81.8 (0.6) & 89.9 (7.2) & 72.6 (4.6) & 82.7 (0.2) & 78.0 (0.5) & 81.0 (2.6) \\ 
\cline{2-9}
 & \multirow{2}{*}{FreqFiT-VPT (ours)} & 3 & 82.6 & 94.2 & 75.8 & 82.2 & 79.4 & 82.8 \\
 &  & 5 & 82.4 (0.3) & 94.0 (0.5)~ & 75.6 (0.1) & 83.1 (0.2) & 79.0 (0.6) & 82.8 (0.3) \\
\bottomrule
\end{tabular}
}
\caption{Per-task fine-tuning results on FGVC tasks with self-supervised learning pre-trained models. We use the same prompt length settings provided in Gated-VPT \citep{vptssl} for all FreqFiT-VPT. We also re-run Gated-VPT using the configuration provided in \citep{vptssl}, including prompt length and learning rate with the same 5 seeds as used for our FreqFiT-VPT, denoted as \textit{Gated-VPT w/ seeds}. The standard deviation from 5 runs is in brackets. Results in \colorbox{mercury}{gray} denote the reported performances in the original Gated-VPT. FreqFiT-VPT denotes the method that our FreqFiT incorporated into VPT. make this a full table if we have space}\label{tab:fgvc}
\end{table}

\begin{figure*}[!t]
\begin{minipage}[]{\linewidth}
    \vskip -3pt
    \rotatebox{90}{\:\:\:Test Acc. (\%)} 
    \hspace{-5pt}
    \includegraphics[width=\linewidth]{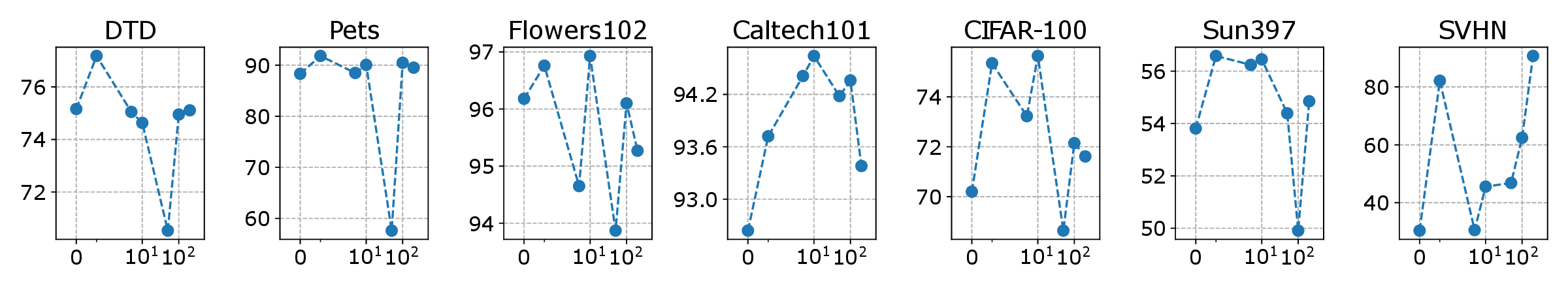}
    \vskip -5pt
    \centering{Prompt length (log scale)}
    \subcaption{FreqFiT-VPT}
\end{minipage}\\
\begin{minipage}[]{\linewidth}
    \rotatebox{90}{\:\:\:Test Acc. (\%)} 
    \hspace{-5pt}
    \includegraphics[width=\textwidth]{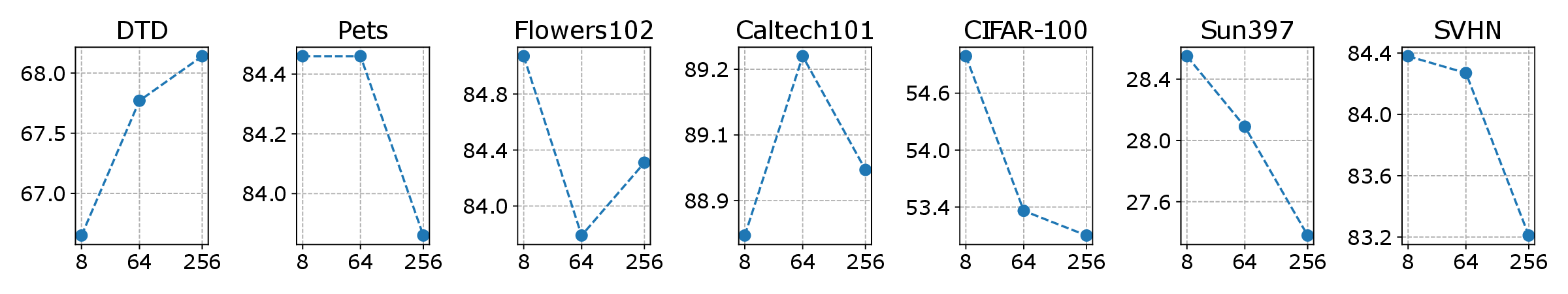}
    \vskip -5pt
    \centering{Reduction factor}
    \subcaption{MoCo / FreqFiT-Adapter}
\end{minipage}
\caption{Ablation on (a) prompt length for FreqFiT-VPT and (b-c) reduction factor for FreqFiT-Adapter on different tasks. We sweep the number of prompts in \{0, 1, 5, 10, 50, 100, 200\} and the reduction factor in \{8, 64, 256\}.}\label{fig:variaty}
\vspace{-10pt}
\end{figure*}

In contrast, the Scaling-Shifting technique offers a straightforward linear transformation by applying learnable scaling and shifting factors to input data, helping to align it with frozen model parameters. This approach adjusts the amplitude and mean of the input features, effectively addressing moderate distribution shifts without modifying the model’s structure. However, Scaling-Shifting’s simplicity limits its ability to capture complex spatial dependencies and high-frequency details, as it focuses only on global feature characteristics. While efficient for tasks with minimal domain shifts, Scaling-Shifting lacks the nuanced adaptability that FreqFit provides. The comparison underscores that while both techniques enhance PEFT performance, FreqFit’s frequency-based modulation better captures intricate data patterns, making it particularly suitable for complex tasks.

\noindent \textbf{Robustness Against Randomization.} Tab. \ref{tab:fgvc} provides per-task fine-tuning results on FGVC tasks with self-supervised learning pre-trained models MAE \citep{mae} and MoCo-v3 \citep{mocov3}. Results in \colorbox{mercury}{gray} denote the reported performances in original Gated-VPT \citep{vptssl}. Since \citep{vptssl} does not provide seeding configurations or how many runs the results were averaged from, we use the provided configurations, including prompt length and learning rate to reproduce their results with the same 5 seeds as used for FreqFiT-VPT, denoted as \textit{Gated-VPT w/ seeds}. Our FreqFiT-VPT demonstrates the stability and superiority over VPT \citep{vpt} and Gated-VPT \citep{vptssl}.
In addition, the results of the \textit{Gated-VPT w/ seed} setting are inconsistent and lower than those in the original paper. We report the standard deviation from 5 runs in brackets. The comparison points out that the \textit{seeding is important} for Gate-VPT and also highlights the stability and improvement of our FreqFiT-VPT.

\textbf{Sensitivity to Hyper-Parameters.} We investigate the degree of performance fluctuation when varying the prompt lengths in VPT and the reduction factor in VPT and Adapter tuning. Fig. \ref{fig:variaty} illustrates the ablation studies on (a) prompt length for FreqFiT-VPT and (b-c) reduction factor for FreqFiT-Adapter across different tasks. We vary the number of prompts in \{0, 1, 5, 10, 50, 100, 200\} and the reduction factor in \{8, 64, 256\}. When adjusting the prompt length, accuracy fluctuates from less than 2\% to roughly 50\% on average for Caltech101 and SVHN tasks, respectively, while other tasks show fluctuations around 5\%. Even with FreqFiT's enhanced ability to capture useful frequency signals, FreqFiT-VPT still heavily depends on the prompt length. In contrast, FreqFiT-Adapter exhibits significantly smaller performance variations, with the largest fluctuation being around 8\% on average. 

\begin{table}[t]
\centering
\resizebox{\columnwidth}{!}{
\begin{tabular}{ccccccccc} 
\toprule
 & \textbf{Attention tuning} & \textbf{DTD} & \textbf{Flowers102} & \textbf{Pets} & \textbf{EuroSAT} & \textbf{Resisc45} & \textbf{Clevr/dist} & \textbf{KITTI/dist} \\ 
\hline
\multirow{2}{*}{\begin{tabular}[c]{@{}c@{}}MAE\end{tabular}} &  & 59.1 & 69.2 & 74.7 & 90.9 & 68.9 & 60.8 & 74.2 \\
 & \checkmark & \textbf{59.5} & \textbf{69.6} & \textbf{75.9} & \textbf{91.4} & \textbf{69.9} & \textbf{61.8} & \textbf{80.3} \\ 
\hline
\multirow{2}{*}{\begin{tabular}[c]{@{}c@{}}MoCo\end{tabular}} &  & 70.2 & 91.0 & 88.6 & 96.5 & 84.7 & 62.5 & 80.7 \\
 & \checkmark & \textbf{70.5} & \textbf{91.0} & \textbf{88.8} & \textbf{96.5} & \textbf{84.7} & \textbf{62.8} & \textbf{81.3} \\
\bottomrule
\end{tabular}
}
\caption{Performance with attention tuning where the FreqFiT module is inserted into the self-attention module. We report the average results from 5 runs.}\label{tab:attn}
\end{table}

\noindent \textbf{Attention Tuning.} In this experiment, we explore an alternative route for learning the long-term dependencies without updating the MSAs. Instead of plugging our frequency-filter module between ViT's blocks, we simply insert it into the self-attention module right after the QKV projection and before the softmax operation. Mathematically:
{\small
\begin{align}
[Q, K, V] &= XW_{q/k/v} + b_{q/k/v} \label{eq:qkv}\\
[Q, K, V] &= \text{FreqFiT}(Q, K, V) \\
\text{Attn}(Q, K, V ) &= \text{Softmax}(QK^T/\sqrt{d})V\label{eq:selfattn1}
\end{align}
}%
where, $W_{q/k/v}$ and $b_{q/k/v}$ are the weights matrices and bias terms of Queue, Key, and Value in the self-attention mechanism. By doing this, we can directly modify the output of the self-attention and learn the long-term dependencies. Tab. \ref{tab:attn} presents the enhanced outcomes of this strategy. Especially, it can boost the results up to 6\% on the KITTI dataset \citep{vtab}. The experiment substantiates the advantage of our frequency-tuning approach. \\

\section{Conclusion}
In this study, we explore the route of frequency adaptation as PEFT method. We introduce FreqFiT, a novel frequency-tuning module designed to modify ViT frequencies for better adaptation to new tasks. This module can be easily integrated into existing fine-tuning methods. We conducted extensive experiments to evaluate FreqFiT on both supervised and self-supervised foundational models, including MAE, MoCo, CLIP, and ImageNet-21k, across VTAB-1k Natural and FGVC tasks. Our comprehensive analysis demonstrates how FreqFiT effectively captures high-frequency components in tokens, leading to significant performance improvements across various tasks. We believe FreqFit establishes a new paradigm for effective adaptation of ViT-based foundation models to downstream tasks.

\small
\bibliographystyle{ieeenat_fullname}
\bibliography{main}

\clearpage
\maketitlesupplementary

\begin{figure*}[t]
\begin{minipage}[]{\linewidth}
\centering{Imagenet-21K - CIFAR100}
\end{minipage}\\
\hspace{10pt}
  \begin{minipage}[]{0.45\linewidth}
    \centering
    \includegraphics[width=0.5\columnwidth]{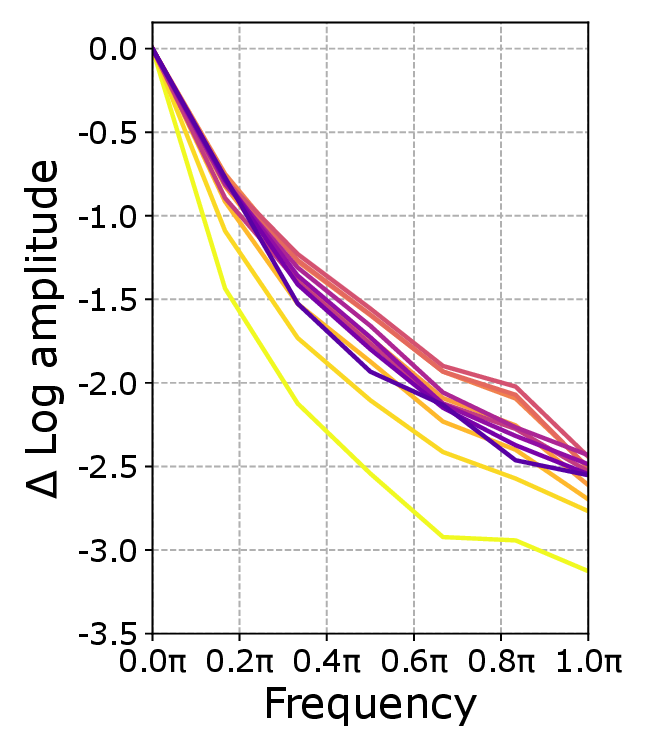} 
    \hspace{-10pt}
    \includegraphics[width=0.5\columnwidth]{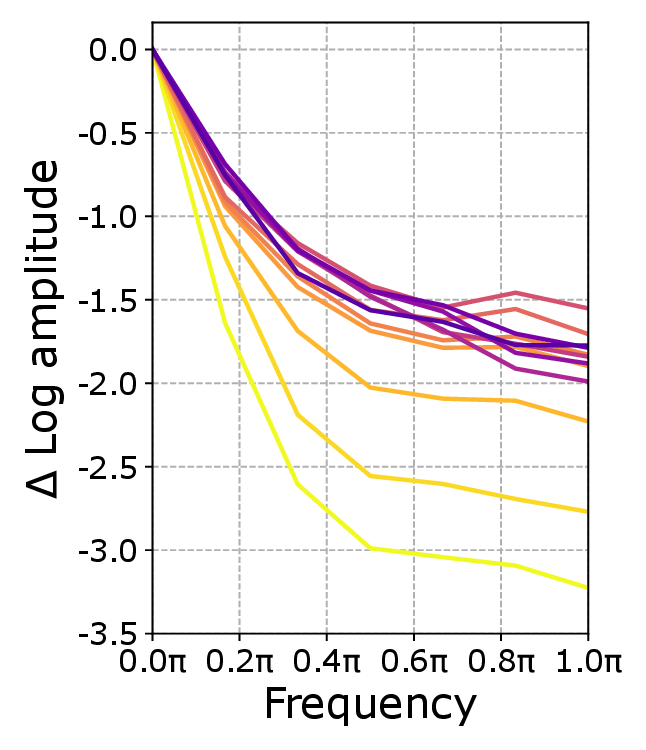}\\    
    \centering{(Left) Bias. (Right) FreqFiT-Bias}
  \end{minipage}
  \begin{minipage}[]{0.1\linewidth}
    \centering
    \includegraphics[width=0.7\columnwidth]{imgs/fig1/heatmap.eps}
  \end{minipage}
    \begin{minipage}[]{0.45\linewidth}
    \centering
    \includegraphics[width=0.5\columnwidth]{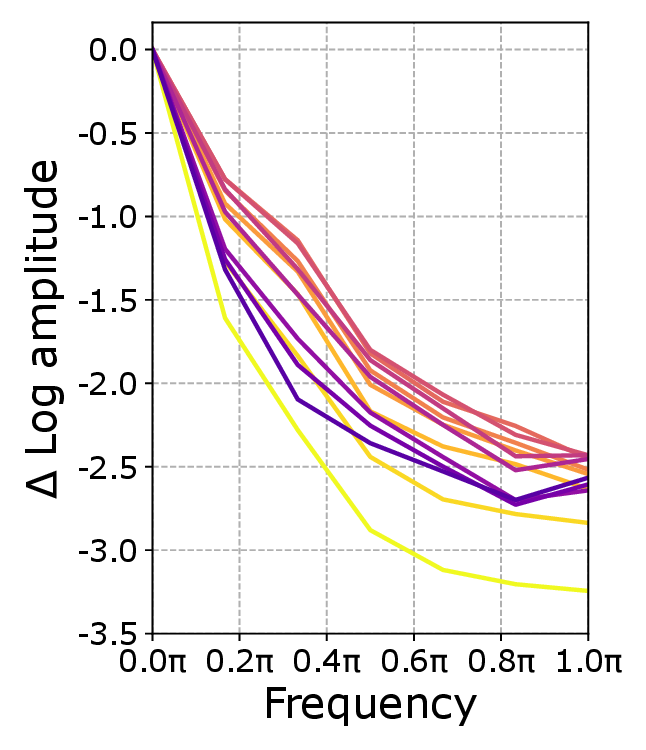} 
    \hspace{-10pt}
    \includegraphics[width=0.5\columnwidth]{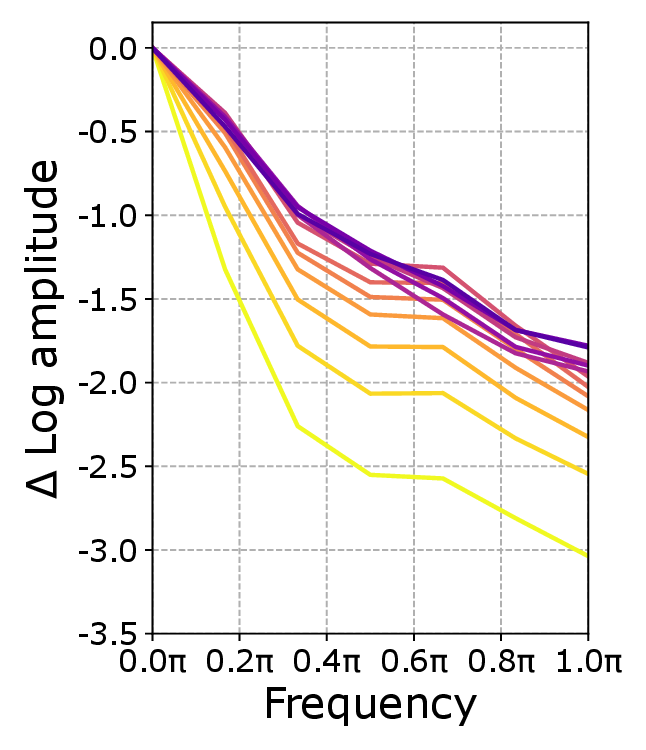}\\   
    \centering{(Left) Adapter. (Right) FreqFiT-Adapter}
  \end{minipage}\\
    \begin{minipage}[]{0.45\linewidth}
    \centering
    \includegraphics[width=0.5\columnwidth]{imgs/features/ori_cifar_238.eps} 
    \hspace{-10pt}
    \includegraphics[width=0.5\columnwidth]{imgs/features/freq_cifar_238.eps}\\    
    \centering{(Left) LoRA. (Right) FreqFiT-LoRA}
  \end{minipage}
  \begin{minipage}[]{0.1\linewidth}
    \centering
    \includegraphics[width=0.7\columnwidth]{imgs/fig1/heatmap.eps}
  \end{minipage}
    \begin{minipage}[]{0.45\linewidth}
    \centering
    \includegraphics[width=0.5\columnwidth]{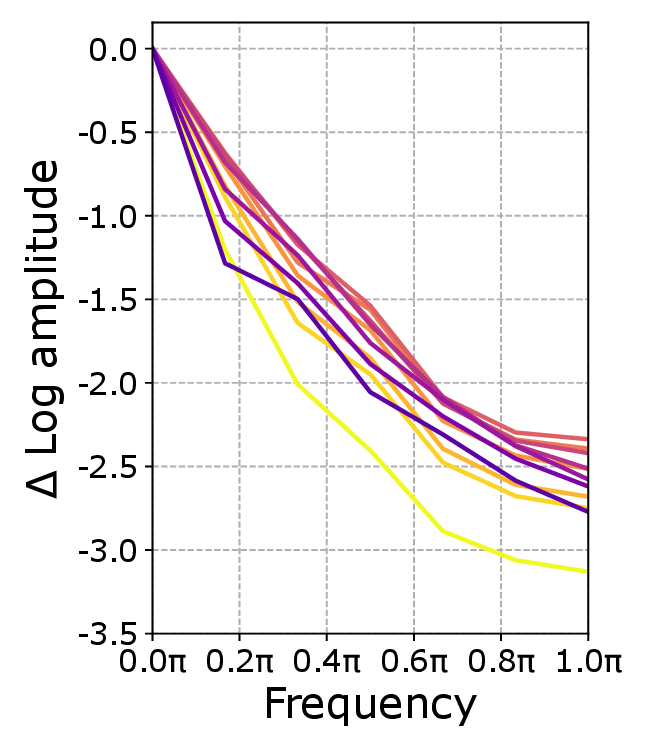} 
    \hspace{-10pt}
    \includegraphics[width=0.5\columnwidth]{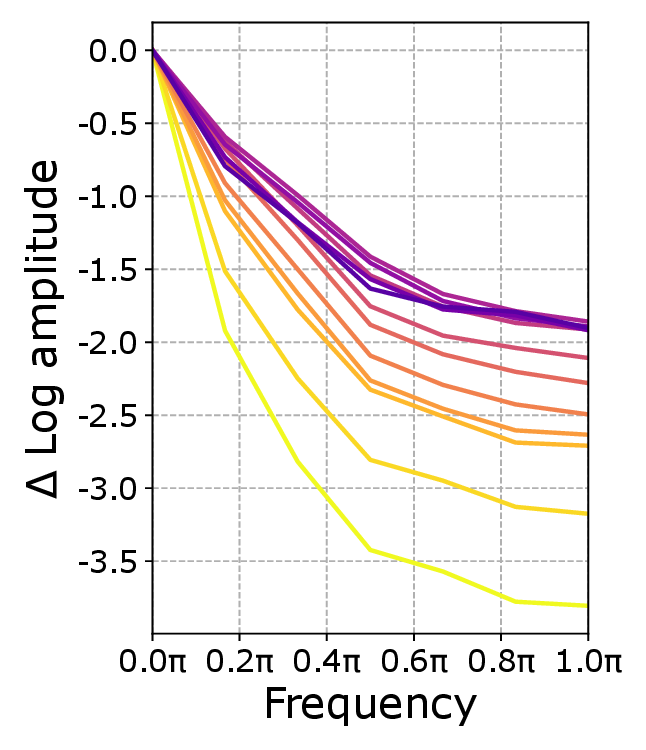}\\   
    \centering{(Left) BOFT. (Right) FreqFiT-BOFT}
  \end{minipage}\\
    \begin{minipage}[]{0.45\linewidth}
    \centering
    \includegraphics[width=0.5\columnwidth]{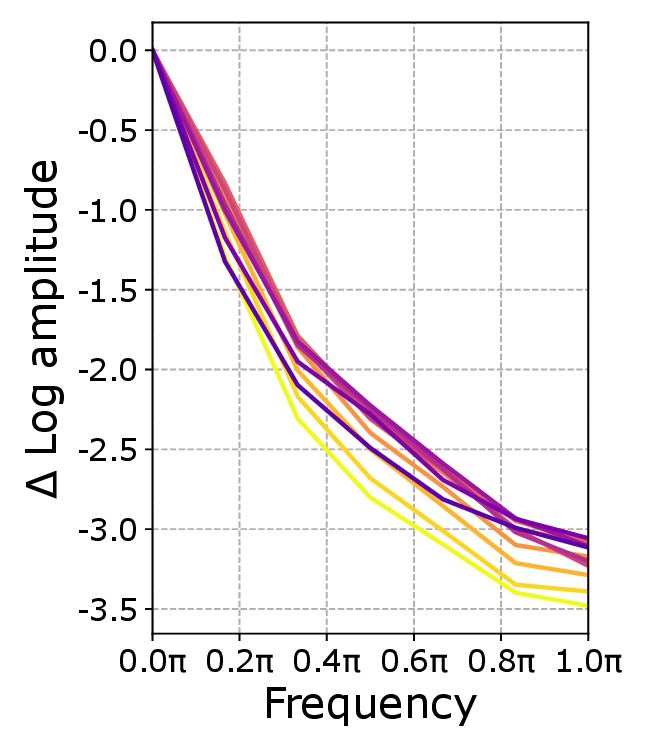} 
    \hspace{-10pt}
    \includegraphics[width=0.5\columnwidth]{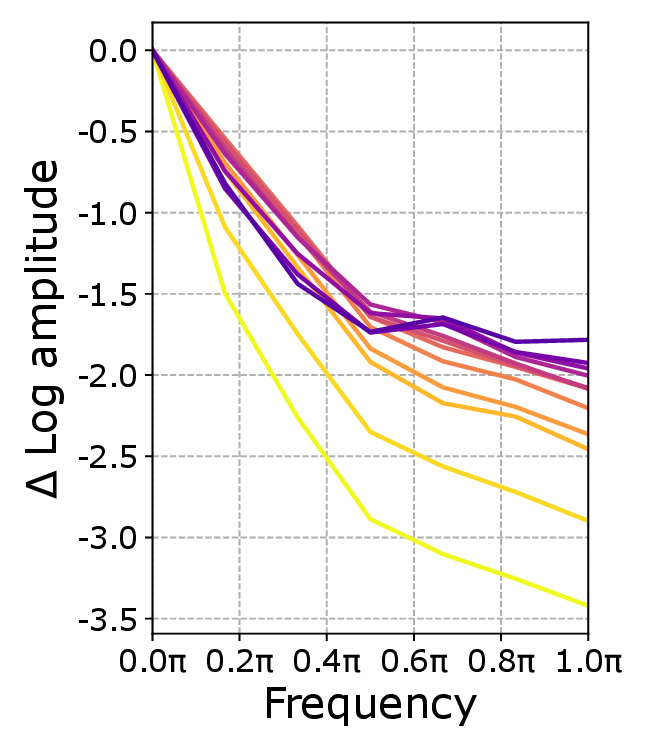}\\    
    \centering{(Left) VeRA. (Right) FreqFiT-VeRA}
  \end{minipage}
  \begin{minipage}[]{0.1\linewidth}
    \centering
    \includegraphics[width=0.7\columnwidth]{imgs/fig1/heatmap.eps}
  \end{minipage}
    \begin{minipage}[]{0.45\linewidth}
    \centering
    \includegraphics[width=0.5\columnwidth]{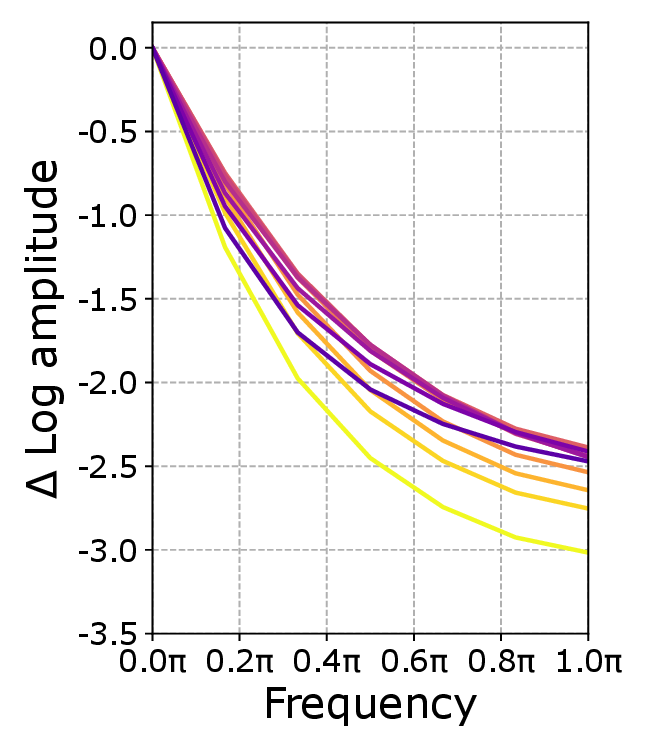} 
    \hspace{-10pt}
    \includegraphics[width=0.5\columnwidth]{imgs/features/imagenet_cifar100_fft_ori_mean.eps}\\   
    \centering{(Left) FourierFT. (Right) FreqFiT-FourierFT}
  \end{minipage}
\caption{Relative log amplitudes of Fourier transformed feature maps of different PEFT methods on Cifar100 with pre-trained Imagenet-21K. $\Delta$ Log amplitude means relative logarithmic amplitude concerning the logarithmic amplitude at normalized frequency $0\pi$ (center) and $1\pi$ (boundary). The brighter the color, the deeper the layer. Our FreqFiT has higher amplitudes compared to the original methods. These visualizations suggest our FreqFit is better at capturing high-frequency components, and potentially leading to better performance.} \label{fig:vizfreq2}
\end{figure*}

The Supplementary Material is organized as follows:
\begin{enumerate}
    \item Proofs for Theorem 1 and Theorem 2, Sec. \ref{sec:proof}.
    \item Augmentation and Hyper-parameters for all experiments, Sec. \ref{sec:hyper}.
    \item FreqFit pseudo-code, Alg. \ref{alg:code}.
    \item Per task result for Scaling Shifting, FreqFit-FourierFT, Tab. \ref{tab:more_vtab21}.
    \item More visualization of the relative log amplitudes of Fourier-transformed feature maps of different PEFT methods, Fig. \ref{fig:vizfreq2}.
    \item More visualization of FreqFit filters on different settings.
\end{enumerate}

\section{Proofs} \label{sec:proof}
Here, we employ LoRA to ease the proof as this can be generalized to other PEFT methods. We re-introduce FreqFit and LoRA equations to facilitate the proofs below. 

Given the input the feature map  $X \in \mathbb{R}^{H \times W \times D}$, where each token is a D-dimensional vector spread across an $H \times W$ spatial grid. The LoRA transformation and FreqFit are as:
\begin{align}
    X_{LoRA} &= W_0X + BAX \label{eq:lora2}\\ 
    X_{FreqFit} &= X + \mathcal{F}^{-1}(\mathcal{F}(X) \odot K) \label{eq:freqfit2}
\end{align}
where, $K \in \mathbb{C}^{H \times W \times D}$, and $B \in \mathbb{R}^{D \times r}$ and $A \in \mathbb{R}^{r \times D}, r  \ll  D$ are the low-rank matrices. Note that, for simplicity, we do not present $\alpha$ and $\beta$ parameters for FreqFit as in the main manuscript. 
\vspace{10pt}

\noindent \underline{\textbf{Theorem 1.}} FreqFit with $O(1)$ parameters can create a feature transformation that spatial-domain parameter-efficient fine-tuning methods cannot replicate.

\noindent \textbf{Proof.} For FreqFit to replicate LoRA transformation, and vice versa, there must exist a filter \textit{F} such that:
\begin{align}
    \mathcal{F}^{-1}(\mathcal{F}(X) \odot K) =  BAX
\end{align}
However, this is not generally possible because of the following reasons:
\begin{enumerate}
    \item Filter \textit{K} and \textit{BA} operate in different domains, i.e., frequency domains and spatial domains, respectively.
    \item \textit{K} is a 3D filter that modulates information in both the tokens 2-dimensional $H \times W$ and the channel dimension \textit{D}. For each position (in frequency domains) in the $H \times W$ grid, \textit{K} contains a unique filter for each of the \textit{D} channels.
    \item \textit{AB}, in Eq. \ref{eq:lora2} is a token-specific modification, where the D-dimensional representation of each token is updated. This modification captures relationships within and across channels, such as correlations or dependencies among the features in D\textit{}.
\end{enumerate}
As a result, FreqFit introduces implicit cross-token interaction via aggregated statistics across all the tokens. Whereas, LoRA operates locally, emphasizing token-specific updates in the channel dimension. 

\textbf{LoRA transformation is not full rank.} The product $BA$, in \ref{eq:lora2}, has a rank of at most \textit{r}, which limits the expressiveness of this transformation to a subspace of dimension \textit{r}. In other words, $BA$ can only capture transformations in an r-dimensional subspace of $\mathbb{R}^{D \times D}$.

\textbf{FreqFit transformation is full rank.}
The Fourier transform of $X \in \mathbb{R}^{H \times W \times D}$ is given by:
{\small
\begin{align}
\hat{\mathbf{X}} &= \mathbf{U}_H \mathbf{X} \mathbf{U}_W^T && \textit{// Fourier transform}\\
\hat{\mathbf{Y}} &= \mathbf{F} \hat{\mathbf{X}} = \mathbf{F} \left( \mathbf{U}_H \mathbf{X} \mathbf{U}_W^T \right) && \textit{// Multiply with F}\\
\mathbf{Y} & = \mathbf{U}_H^{H} \hat{\mathbf{Y}} (\mathbf{U}_W^{H})^T && \textit{// Inverse Fourier transform}
\end{align}
}%
where $\mathbf{U}_H \in \mathbb{C}^{H \times H}$ and $\mathbf{U}_W \in \mathbb{C}^{W \times W}$ are the unitary Fourier transform matrix of row and column grid. The $\mathbf{U}^H_H$ and $\mathbf{U}^H_W$ are the conjugate transpose of $\mathbf{U}_H$ and $\mathbf{U}_W$, respectively, i.e., $\mathbf{U}^H_H \mathbf{U}_H = \mathbf{I}$ and $\mathbf{U}^H_W \mathbf{U}_W = \mathbf{I}$. Thus, if the input $\mathbf{X}$ is full-rank and $\mathbf{F}$ is a full-rank diagonal matrix (no zero entries for all frequency components), since $\mathbf{U}$  is unitary, the rank of $\mathbf{X}$ is preserved in the frequency domain, then the result $\mathbf{X}$ must be full-rank. This means FreqFit can learn from all information from the input feature with $O(1)$.

\textbf{FreqFit has $O(1)$ parameter complexity}, as the frequency modulation filter $F$ can be parameterized efficiently, focusing only on essential frequency components, regardless the input dimensions of \textit{X}.

Therefore, FreqFit with $O(1)$ parameters can create a feature transformation that spatial-domain parameter-efficient fine-tuning methods cannot replicate. 

\hfill$\square$ 

\vspace{10pt}
\noindent \underline{\textbf{Theorem 2.}} Combining FreqFit with spatial-domain PEFT methods can create a feature transformation that cannot be achieved by FreqFit or any spatial-domain PEFT method alone.

\noindent \textbf{Proof.} The FreqFit filter $K \in \mathbb{C}^{H \times W \times D}$ applies a frequency-domain filter independently to each channel $d \in D$ on the grid $H \times W$. This implies $X_{FreqFit}[h, w, d]$ depends only on $X_{FreqFit}[h', w', d]$ as in Eq. \ref{eq:freqfit3}. Whereas, the product $BA$, Eq. \ref{eq:lora2} is the same across all tokens, meaning it only introduces channel-wise dependencies. This means $X_{LoRA}[h, w, d]$ depends only on $X_{LoRA}[h, w, d']$ as in Eq. \ref{eq:lora3}. Mathematically:
{\small
\begin{align}
    X_{FreqFiT}[h, w, d]\: &\parallel  \:X_{FreqFit}[h', w', d]\:\:\: \forall h', w' \label{eq:freqfit3} \\
    X_{LoRA}[h, w, d]\: &\parallel  \:X_{LoRA}[h, w, d'] \label{eq:lora3}
\end{align}
}%
Assume FreqFit can replicate LoRA transformation. Then, FreqFit must create cross-channel dependencies of the form
{\small
\begin{align}
    X_{FreqFiT}[h, w, d]\: &\parallel  \:X_{FreqFit}[h', w', d']\:\:\: \text{for   } d \neq d'
\end{align}
}%
However, by construction, FreqFit only operates independently within each channel $d \in D$. This directly contradicts LoRA as it introduces dependencies across $d \in D$. Thus, \textit{FreqFit can not replicate LoRA transformation}.

Together with the result of Theorem 1 which shows LoRA can not replicate FreqFit transformation, Theorem 2 provides a compelling rationale for combining two complementary approaches: FreqFit and PEFT methods like LoRa. By leveraging their distinct strengths, combining these two methods can yield transformations that neither method can achieve on its own.

\hfill$\square$

\section{Augmentation and Hyper-parameters}\label{sec:hyper}

We use PyTorch to implement all experiments on NVIDIA V100-32GB GPUs. Following \cite{vpt}, we conduct a grid search to find the tuning-specific hyper-parameters, learning rate, and weight decay values using val set of each task, as shown in Tab. \ref{tab:setting}.

\begin{table*}[h]
\caption{Implementation details for all of our experiments.}\label{tab:setting} 
\centering
\resizebox{\textwidth}{!}{
\begin{tabular}{lcc} 
\toprule
 & PEFT & FreqFiT \\
\hline
Optimizer & AdamW & SGD \\
Optimizer momentum & - & 0.9 \\
Batch size& \multicolumn{2}{c}{64} \\
Image size& \multicolumn{2}{c}{224 $\times$ 224} \\
Base lr range~ ~ & \{0.2, 0.1, 0.05, 0.02, 0.01, 0.005, 0.001, 0.0001, 0.0002, 0.0005\} & \{10., 5., 2.5, 1.,0.5, 0.25, 0.1, 0.05, 0.025, 0.005\} \\
Weight decay range & \multicolumn{2}{c}{\{0.01, 0.001, 0.0001, 0.0\}} \\
Learning rate schedule & \multicolumn{2}{c}{cosine decay~~} \\
Warm up epochs & \multicolumn{2}{c}{10} \\
Total epochs & \multicolumn{2}{c}{100} \\
Augmentation & \multicolumn{2}{c}{Follow  as in VPT \cite{vpt}} \\
Layer adaptation & \multicolumn{2}{c}{LoRA, VeRA, BOFT, and FourierFT experiments, all Linear layer are adapted} \\
Low-rank configuration & \multicolumn{2}{c}{LoRA, VeRA, BOFT, and FourierFT use default configuration from HuggingFace \cite{huggingface}} \\
Prompt length & \multicolumn{2}{c}{MAE follows Gated-VPT \cite{vptssl}; MoCo and Imagenet-21K follow VPT \cite{vpt}} \\
\bottomrule
\end{tabular}
}
\end{table*}

\begin{algorithm*}[h]
\caption{Implementation of FreqFiT in PyTorch-like style.}\label{alg:code}
\definecolor{codeblue}{rgb}{0.25,0.5,0.5}
\definecolor{colorred}{RGB}{197, 49, 124}
\lstset{
  backgroundcolor=\color{white},
  basicstyle=\fontsize{6.8pt}{6.8pt}\ttfamily\selectfont,
  columns=fullflexible,
  breaklines=true,
  captionpos=b,
  commentstyle=\fontsize{6.8pt}{6.8pt}\color{codeblue},
  keywordstyle=\fontsize{6.8pt}{6.8pt}\color{colorred},
}
\begin{minipage}[]{\textwidth}
\begin{lstlisting}[language=python]
class FreqFiT(nn.Module):
    def __init__(self, dim, h=14, w=8):
        # dim: token dimension
        # filter_weight: H_hat x D, where H_hat = (H x W) // 2 + 1
        super().__init__()
        # initialize filter weights
        self.filter_weight = nn.Parameter(torch.randn(h, w, dim, 2, dtype=torch.float32) * 0.02)
        self.scale = nn.Parameter(torch.ones(dim))
        self.shift = nn.Parameter(torch.zeros(dim))
    
        nn.init.normal_(self.scale, mean=1, std=.02)
        nn.init.normal_(self.shift, std=.02)
        
    def forward(self, x):
        B, N, C = x.shape
        x = x.view(B, a, b, C).to(torch.float32)
        res = x
        x = torch.fft.rfft2(x, dim=(1, 2), norm='ortho')
        weight = torch.view_as_complex(self.complex_weight.squeeze())
        x = x * weight
        x = torch.fft.irfft2(x, s=(a, b), dim=(1, 2), norm='ortho')
        x = x * self.scale + self.shift
        x = x + res
        x = x.reshape(B, N, C)
        return x
\end{lstlisting}
\end{minipage}
\end{algorithm*}

\begin{table*}[t]
\centering
\resizebox{\textwidth}{!}{
\begin{tabular}{cl|c|ccccccc|cccc|ccccccccc} 
\toprule
 &  &  & \multicolumn{7}{c|}{Natural} & \multicolumn{4}{c|}{Specialized} & \multicolumn{8}{c}{Structured} &  \\
\vcell{} & \multicolumn{1}{c|}{\vcell{ViT-B/16}} & \vcell{\begin{tabular}[b]{@{}c@{}}NO\\Mixup/\\Strong\\Aug.\end{tabular}} & \vcell{\begin{sideways}Cifar100\end{sideways}} & \vcell{\begin{sideways}Caltech101\end{sideways}} & \vcell{\begin{sideways}DTD\end{sideways}} & \vcell{\begin{sideways}Flower102\end{sideways}} & \vcell{\begin{sideways}Pets\end{sideways}} & \vcell{\begin{sideways}SVHN\end{sideways}} & \vcell{\begin{sideways}Sun397\end{sideways}} & \vcell{\begin{sideways}Camelyon\end{sideways}} & \vcell{\begin{sideways}EuroSAT\end{sideways}} & \vcell{\begin{sideways}Resisc45\end{sideways}} & \vcell{\begin{sideways}Retinopathy\end{sideways}} & \vcell{\begin{sideways}Clevr-Count\end{sideways}} & \vcell{\begin{sideways}Clevr-Dist\end{sideways}} & \vcell{\begin{sideways}DMLab\end{sideways}} & \vcell{\begin{sideways}KITTI-Dist\end{sideways}} & \vcell{\begin{sideways}dSpr-Loc\end{sideways}} & \vcell{\begin{sideways}dSpr-Ori\end{sideways}} & \vcell{\begin{sideways}sNORB-Azim\end{sideways}} & \vcell{\begin{sideways}sNORB-Elev\end{sideways}} & \vcell{\begin{sideways}Mean\end{sideways}} \\[-\rowheight]
\printcellmiddle & \multicolumn{1}{c|}{\printcellmiddle} & \printcellmiddle & \printcellbottom & \printcellbottom & \printcellbottom & \printcellbottom & \printcellbottom & \printcellbottom & \printcellbottom & \printcellbottom & \printcellbottom & \printcellbottom & \printcellbottom & \printcellbottom & \printcellbottom & \printcellbottom & \printcellbottom & \printcellbottom & \printcellbottom & \printcellbottom & \printcellbottom & \printcellbottom \\ 
\hline
 & Full & - & 68.9 & 87.7 & 64.3 & 97.2 & 86.9 & 87.4 & 38.8 & 79.7 & 95.7 & 84.2 & 73.9 & 56.3 & 58.6 & 41.7 & 65.5 & 57.5 & 46.7 & 25.7 & 29.1 & 68.9 \\ 
\hline
 & Linear & \cmark & 64.4 & 85.0 & 63.2 & 97.0 & 86.3 & 36.6 & 51.1 & 78.5 & 87.5 & 68.5 & 74.0 & 34.3 & 30.6 & 33.2 & 55.4 & 12.5 & 20.0 & 9.6 & 19.2 & 57.6 \\
 & Bias & \cmark & 72.8 & 87.0 & 59.2 & 97.5 & 85.3 & 59.9 & 51.4 & 78.7 & 91.6 & 72.9 & 69.8 & 61.5 & 55.6 & 32.4 & 55.9 & 66.6 & 40.0 & 15.7 & 25.1 & 65.2 \\
 & Adapter-64 & \cmark & 74.2 & 85.8 & 62.7 & 97.6 & 87.2 & 36.3 & 50.9 & 76.3 & 87.5 & 73.7 & 70.9 & 42.9 & 39.9 & 30.4 & 54.5 & 31.9 & 25.6 & 13.5 & 21.4 & 56.0 \\
 & VPT & \cmark & 78.8 & 90.8 & 65.8 & 98.0 & 88.3 & 78.1 & 49.6 & 81.8 & 96.1 & 83.4 & 68.4 & 68.5 & 60.0 & 46.5 & 72.8 & 73.6 & 47.9 & 32.9 & 37.8 & 72.0 \\
 & LoRa & \cmark & 82.9 & 90.8 & 66.9 & 98.7 & 89.9 & 83.8 & 55.1 & 84.5 & 95.7 & 83.4 & 74.7 & 72.2 & 58.5 & 47.8 & 75.4 & 75.7 & 47.1 & 23.0 & 28.1 & 73.1 \\
 & BOFT & \cmark & 78.7 & 90.5 & 68.4 & 98.5 & 88.3 & 83.2 & 53.5 & 81.2 & 95.9 & 82.7 & 70.3 & 72.2 & 60.4 & 37.3 & 74.1 & 63.9 & 42.9 & 26.0 & 30.6 & 71.2 \\
 & VeRA & \cmark & 80.7 & 90.7 & 68.9 & 98.5 & 89.5 & 84.9 & 53.5 & 81.3 & 95.2 & 83.2 & 73.4 & 73.6 & 60.3 & 43.6 & 77.2 & 72.7 & 46.2 & 28.5 & 30.3 & 72.8 \\
& FourierFT & \cmark & 79.8 & 90.4 & 68.8 & 98.3 & 89.9 & 85.0 & 53.1 & 82.7 & 95.2  & 82.6 & 74.9 & 73.5 & 60.1 & 43.0  & 76.9 & 71.2 & 47.6 & 28.1  & 29.4  & 72.8 \\
 & SSF & \xmark & 69.0 & 92.6 & 75.1 & 99.4 & 91.8 & 90.2 & 52.9 & 87.4 & 95.9 & 87.4 & 75.5 & 75.9 & 62.3 & 53.3 & 80.6 & 77.3 & 54.9 & 29.5 & 37.9 & 75.7 \\
 & NOAH & \xmark & 69.6 & 92.7 & 70.2 & 99.1 & 90.4 & 86.1 & 53.7 & 84.4 & 95.4 & 83.9 & 75.8 & 82.8 & 68.9 & 49.9 & 81.7 & 81.8 & 48.3 & 32.8 & 44.2 & 75.5 \\
 & AdaptFormer & \xmark & 70.8 & 91.2 & 70.5 & 99.1 & 90.9 & 86.6 & 54.8 & 83.0 & 95.8 & 84.4 & 76.3 & 81.9 & 64.3 & 49.3 & 80.3 & 76.3 & 45.7 & 31.7 & 41.1 & 74.7 \\
 & RepAdaptor & \xmark & 72.4 & 91.6 & 71.0 & 99.2 & 91.4 & 90.7 & 55.1 & 85.3 & 95.9 & 84.6 & 75.9 & 82.3 & 68.0 & 50.4 & 79.9 & 80.4 & 49.2 & 38.6 & 41.0 & 76.1 \\ 
\hline
\multirow{6}{*}{\begin{tabular}[c]{@{}c@{}}scale-shift\\(ours)\end{tabular}} & Linear & \multirow{6}{*}{\cmark} & \textbf{77.8} & \textbf{89.8} & \textbf{66.5} & \textbf{98.5} & \textbf{88.1} & \textbf{75.9} & \textbf{53.8} & 75.7 & \textbf{92.3} & \textbf{79.4} & 71.2 & \textbf{71.7} & \textbf{60.1} & \textbf{37.1} & \textbf{70.4} & \textbf{53.9} & \textbf{37.4} & \textbf{25.7} & \textbf{34.0} &  \textbf{69.0} (\textcolor{green}{+11.4})\\
 & Bias &  & \textbf{80.1} & \textbf{91.2} & \textbf{68.7} & \textbf{98.7} & \textbf{88.7} & \textbf{75.9} & \textbf{54.7} & \textbf{79.7} &  \textbf{95.4} & \textbf{82.5} & \textbf{70.1} & \textbf{69.3} & \textbf{61.2} & \textbf{39.6} & \textbf{70.0} &  \textbf{71.5} & \textbf{47.4}  & \textbf{24.5} & \textbf{31.2} & \textbf{71.2} (\textcolor{green}{+6.0}) \\
 & Adapter-64 &  & \textbf{79.8} & \textbf{89.2} & \textbf{68.0} & \textbf{98.6} & \textbf{89.2} & \textbf{77.2} & \textbf{54.5} & \textbf{82.7} & \textbf{95.4} & \textbf{82.8} & \textbf{73.5} & \textbf{70.4} & \textbf{58.1} & \textbf{36.0} & \textbf{62.5} & \textbf{62.0} & \textbf{41.6} & \textbf{25.8} & \textbf{37.9} & \textbf{70.8} (\textcolor{green}{+14.8})\\
 & VPT &  & \textbf{79.8} & \textbf{90.8} & \textbf{67.5} & \textbf{98.5} & 87.6 & \textbf{79.2} & \textbf{53.9} & \textbf{83.1} & 94.4 & 79.6 & \textbf{72.7} & \textbf{73.3} & \textbf{61.1} & 45.0 & \textbf{74.4} & \textbf{78.8} & \textbf{50.1} & 32.7 & 37.3 & \textbf{72.9 }(\textcolor{green}{+0.9})\\
 & LoRa &  & 77.3 & 89.7 & \textbf{67.1} & 98.4 & 88.9 & \textbf{86.0} & 53.8 & \textbf{85.5} & 94.8 & 78.5 & 73.6 & 71.7 & 55.7 & 42.8 & \textbf{77.6} & 66.6 & 43.9 & \textbf{31.5} & \textbf{33.4} & 72.1 (-1.0) \\
 & BOFT &  & 78.6 & \textbf{91.0} & 67.5 & \textbf{98.7} & \textbf{88.6} & \textbf{85.9} & \textbf{53.6} & 79.4 & \textbf{95.9} & \textbf{84.1} & \textbf{73.3} & \textbf{73.8} & \textbf{60.8} & \textbf{38.4} & \textbf{76.0} & \textbf{65.6} & \textbf{45.0} & \textbf{31.5} & \textbf{32.4} & \textbf{72.2} (\textcolor{green}{+1.0})\\ 
\hline
\multirow{7}{*}{\begin{tabular}[c]{@{}c@{}}FreqFit-\\(ours)\end{tabular}} & Linear & \multirow{7}{*}{\cmark} & \textbf{78.5} & \textbf{89.8} & \textbf{70.1} & \textbf{98.6} & \textbf{88.7} & \textbf{76.0} & \textbf{53.9} & \textbf{79.3} & \textbf{94.8} & \textbf{81.5} & 72.9 & \textbf{72.0} & \textbf{60.1} & \textbf{39.6} & \textbf{70.5} & \textbf{71.2} & \textbf{39.0} & \textbf{33.0} & \textbf{34.5} &  \textbf{71.3} (\textcolor{green}{+13.7})\\
 & Bias &  & \textbf{82.4} & \textbf{90.4} & \textbf{69.0} & \textbf{98.9} & \textbf{89.5} & \textbf{78.7} & \textbf{54.9} & \textbf{79.9} & \textbf{95.9} & \textbf{82.0} & \textbf{70.3} & \textbf{64.9} & \textbf{59.7} & \textbf{35.7} & \textbf{72.4} & \textbf{74.8} & \textbf{41.4} & \textbf{24.7} & \textbf{36.3} & \textbf{71.3}
 (\textcolor{green}{+6.1}) \\
 & Adapter-64 &  & \textbf{81.5} & \textbf{90.4} & \textbf{70.5} & \textbf{98.7} & \textbf{89.3} & \textbf{82.3} & \textbf{54.2} & \textbf{81.7} & \textbf{96.1} & \textbf{82.3} & \textbf{73.4} & \textbf{71.4} & 61.2 & \textbf{37.2} & \textbf{63.6} & \textbf{84.2} & \textbf{40.1} & \textbf{26.4} & \textbf{38.7} & \textbf{72.4} (\textcolor{green}{+16.4})\\
 & VPT &  & \textbf{80.1} & \textbf{92.8} & \textbf{68.4} & \textbf{98.5} & \textbf{89.1} & \textbf{86.2} & \textbf{54.6} & \textbf{84.4} & \textbf{96.1} & \textbf{83.9} & \textbf{73.1} & \textbf{73.8} & 59.2 & 45.6 & \textbf{75.8} & \textbf{75.4} & \textbf{47.9} & \textbf{35.2} & \textbf{44.8} & \textbf{74.3} (\textcolor{green}{+2.4})\\
 & LoRa &  & \textbf{83.7} & 90.2 & \textbf{68.8} & \textbf{98.9} & \textbf{90.5} & \textbf{86.3} & \textbf{55.1} & \textbf{85.0} & \textbf{96.1} & \textbf{83.5} & \textbf{74.7} & \textbf{74.5} & \textbf{62.0} & 45.9 & \textbf{78.2} & \textbf{78.7} & \textbf{51.7} & \textbf{31.7} & \textbf{36.1} & \textbf{74.7} (\textcolor{green}{+1.6}) \\
 & BOFT &  & 78.1 & \textbf{91.0} & 68.0 & \textbf{98.6} & \textbf{89.5} & \textbf{86.0} & \textbf{54.3} & 80.3 & \textbf{96.1} & \textbf{83.4} & \textbf{74.0} & \textbf{73.8} & \textbf{60.8} & \textbf{38.2} & \textbf{78.2} & \textbf{73.7} & \textbf{45.0} & \textbf{30.0} & \textbf{33.0  } & \textbf{72.7} (\textcolor{green}{+1.5}) \\
& VeRA &  & \textbf{81.3} & 90.4 & \textbf{71.6} & \textbf{98.9} & \textbf{90.3 }& \textbf{85.7} & \textbf{54.6} & 80.6 & \textbf{95.9} & \textbf{83.7} & \textbf{73.4} & 69.4 & \textbf{60.5} & 42.5 & \textbf{77.6} & \textbf{83.1} & 44.0  & \textbf{32.0} & \textbf{33.6} &  \textbf{73.5} (\textcolor{green}{+0.8})\\
& FourierFT &  & \textbf{83.8} & \textbf{91.7} & \textbf{72.1} & \textbf{98.9} & \textbf{90.5} & \textbf{87.5} & \textbf{54.2} & \textbf{84.0} & \textbf{95.2} & \textbf{83.6} & 73.6 & 73.3 & 59.8 & \textbf{46.0} & 76.1 & \textbf{76.8} & \textbf{50.7} &\textbf{30.4}& \textbf{33.3} &  \textbf{74.2} (\textcolor{green}{+1.4}) \\
\bottomrule
\end{tabular}
}
\caption{Comparison between tuning the data with scale-shift and frequency tuning method and their counterpart state-of-the-arts on VTab dataset with Imagenet-21K pre-trained weights. \textbf{Bold} indicates that the performance is better than the original method without tuning the data. Compared to Tab 1 in the main manuscript, we add the performances of all tasks that are used to produce Fig. 4 in the main manuscript. We also add the FourierFT \cite{fourierft} baseline and \textit{FreqFit-FourierFT}.
}\label{tab:more_vtab21}
\end{table*}

\begin{figure*}[h]
\centering
\begin{minipage}[]{0.5\linewidth}
    \centering{ImageNet / FreqFiT-LoRA / Cifar100}
    \includegraphics[width=\textwidth]{imgs/filters/imagenet_lora_filter.eps} 
    \centering{(a) Mean of filters from 12 layers. \\Left to right, Upper row: layer 1-6. Lower row: layer 7-12}
\end{minipage}\\
\vspace{10pt}
\begin{minipage}[]{0.8\linewidth}
    \includegraphics[trim=80 15 80 15,clip,width=\textwidth]{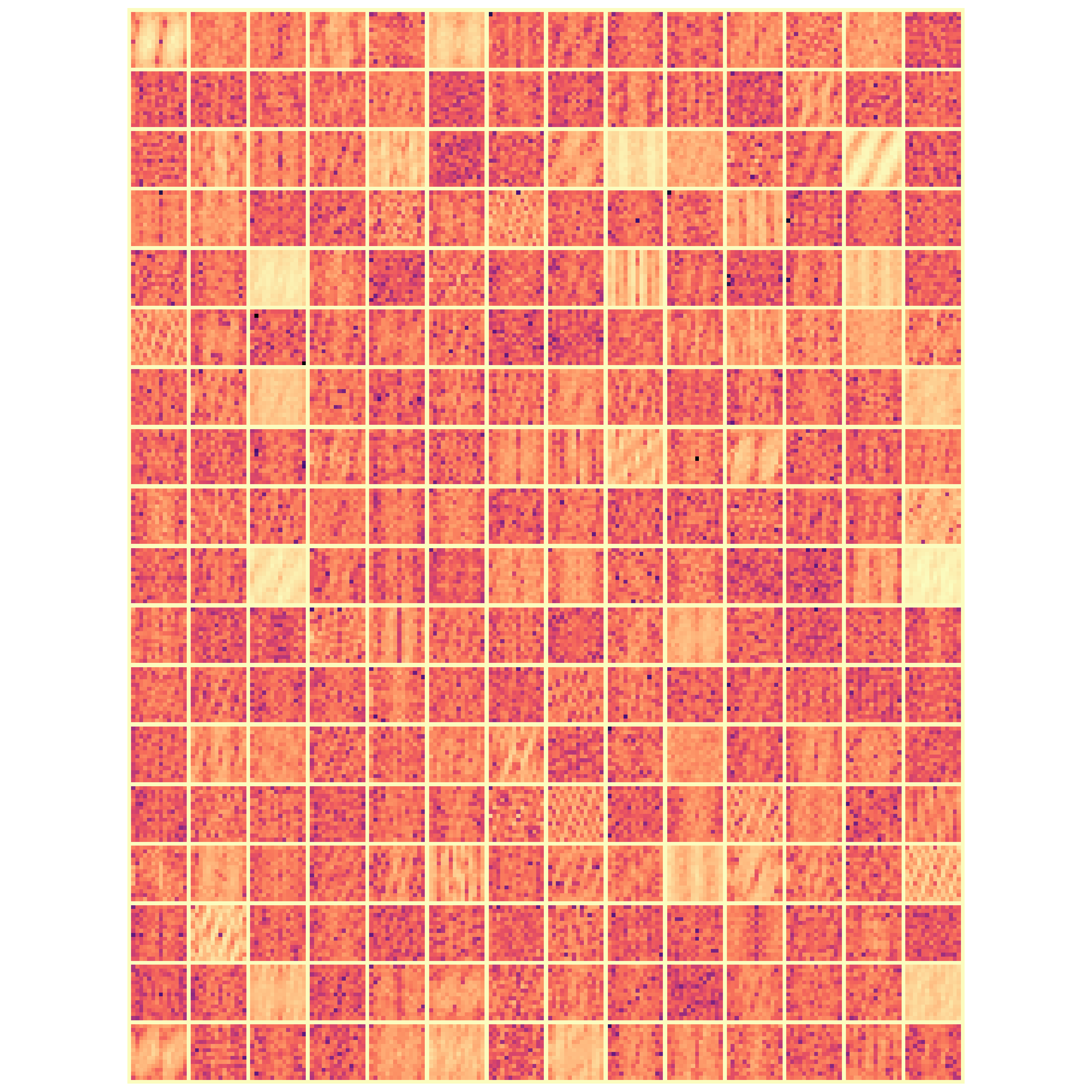} 
    \centering{(b) First 252 filters from the 1st layer}
\end{minipage}
\caption{(a) Mean of filters from 12 layers from the setting ImageNet / FreqFiT-LoRA / Cifar100, presented in Figure 3 in the main paper. (b) First 252 filters from the 1st layer of the same setting.
} \label{fig:filter_mae}
\end{figure*}

\begin{figure*}[h]
\centering
\begin{minipage}[]{0.5\linewidth}
    \centering{ImageNet / FreqFiT-FourierFT / Cifar100}
    \includegraphics[width=\textwidth]{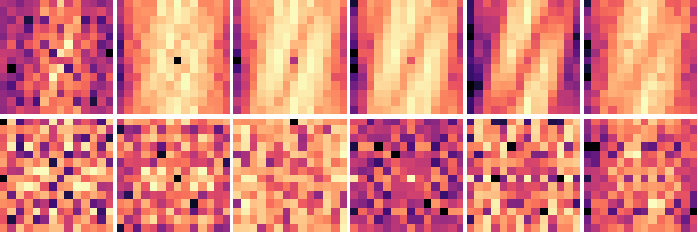} 
    \centering{(a) Mean of filters from 12 layers. \\Left to right, Upper row: layer 1-6. Lower row: layer 7-12}
\end{minipage}\\
\vspace{10pt}
\begin{minipage}[]{0.8\linewidth}
    \includegraphics[trim=80 15 80 15,clip,width=\textwidth]{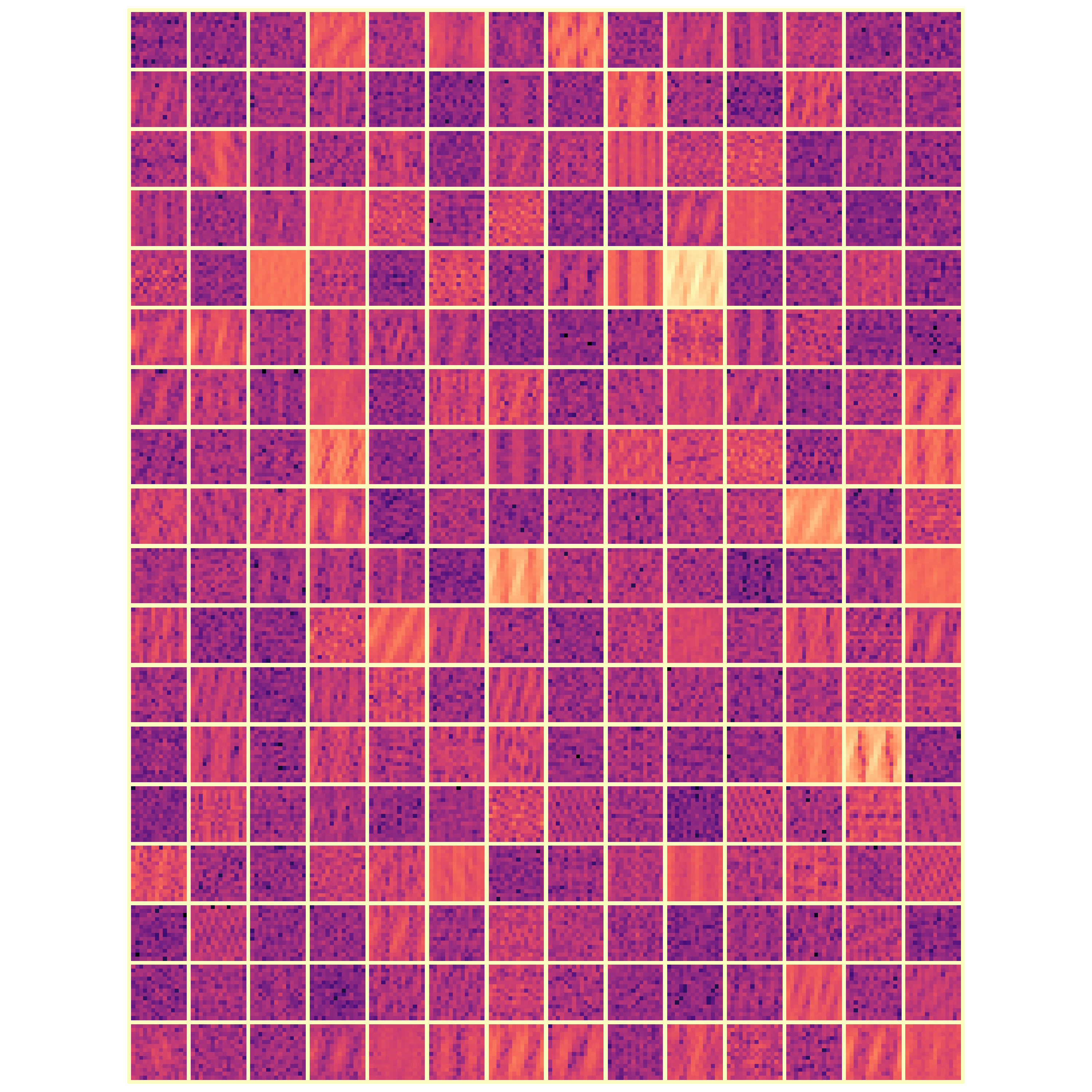} 
    \centering{(b) First 252 filters from the 1st layer}
\end{minipage}
\caption{(a) Mean of filters from 12 layers from the setting ImageNet / FreqFiT-FourierFT / Cifar100. (b) First 252 filters from the 1st layer of the same setting.
} \label{fig:filter_mae}
\end{figure*}

\begin{figure*}[h]
\centering
\begin{minipage}[]{0.5\linewidth}
    \centering{MAE / FreqFiT-Adapter / Caltech101}
    \includegraphics[trim=55 15 40 15,clip,width=\textwidth]{imgs/filters/mae_adapter_vtab_cal.eps} 
    \centering{(a) Mean of filters from 12 layers. \\Left to right, Upper row: layer 1-6. Lower row: layer 7-12}
\end{minipage}\\
\vspace{10pt}
\begin{minipage}[]{0.8\linewidth}
    \includegraphics[trim=80 15 80 15,clip,width=\textwidth]{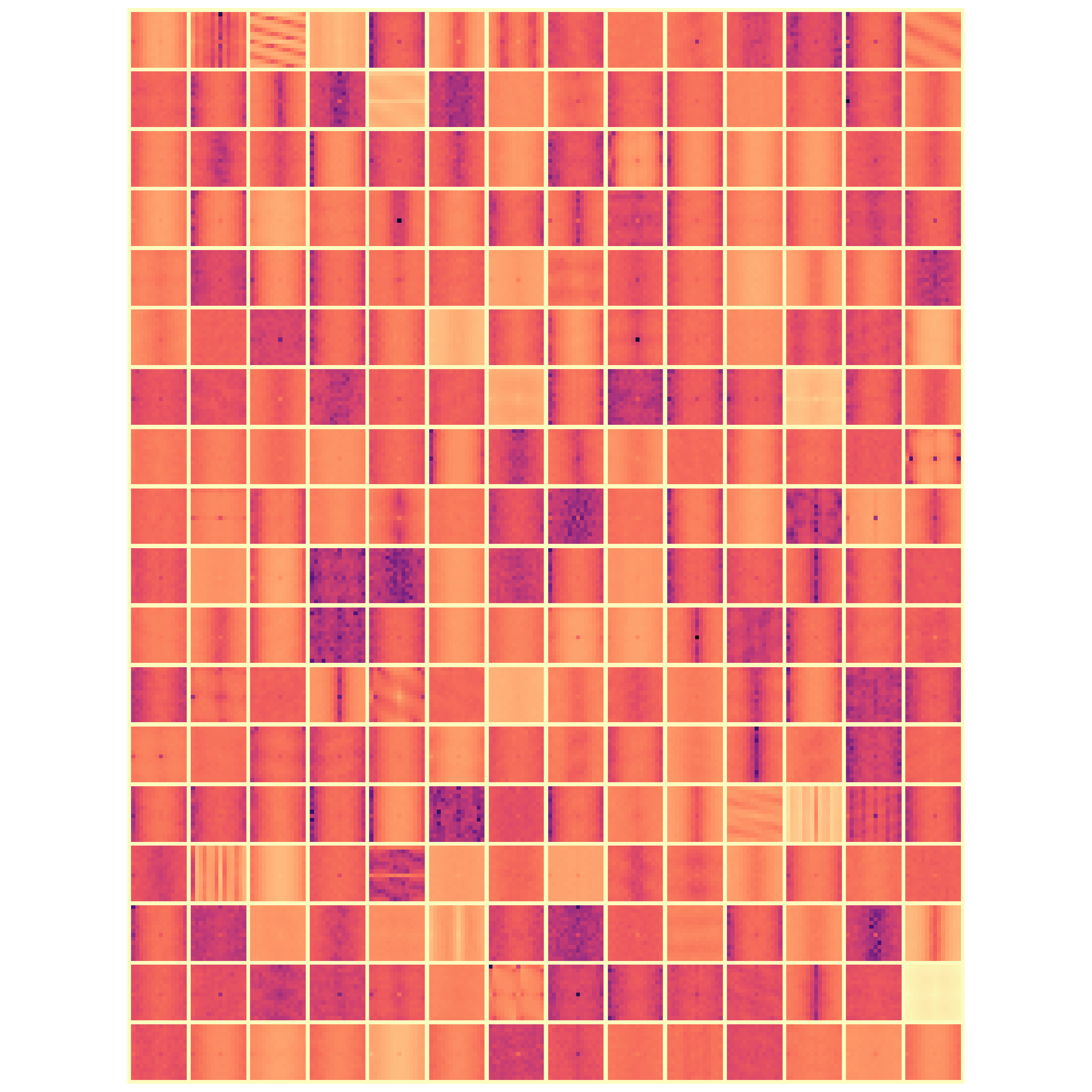} 
    \centering{(b) First 252 filters from the last layer}
\end{minipage}
\caption{(a) Mean of filters from 12 layers from the setting MAE / FreqFiT-Adapter / Caltech101, presented in Figure 3 in the main paper. (b) First 252 filters from the last layer of the same setting.
} \label{fig:filter_mae}
\end{figure*}

\begin{figure*}[h]
\centering
\begin{minipage}[]{0.5\linewidth}
    \centering{MAE / FreqFiT-Bias / Caltech101}
    \includegraphics[trim=55 15 40 15,clip,width=\textwidth]{imgs/filters/mae_bias_vtab_cal.eps} 
    \centering{(a) Mean of filters from 12 layers. \\Left to right, Upper row: layer 1-6. Lower row: layer 7-12}
\end{minipage}\\
\vspace{10pt}
\begin{minipage}[]{0.8\linewidth}
    \includegraphics[trim=80 15 80 15,clip,width=\textwidth]{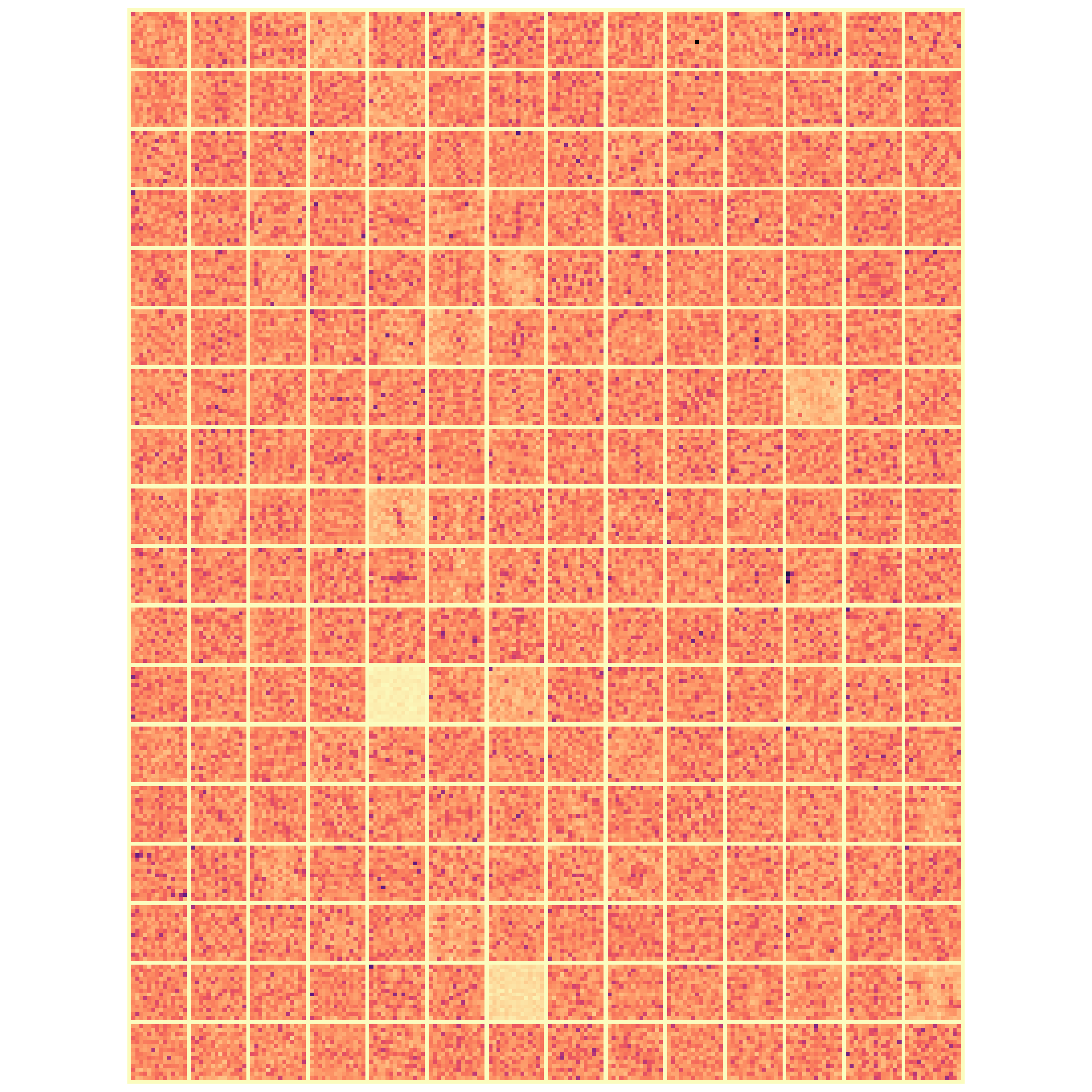} 
    \centering{(b) First 252 filters from the last layer.}
\end{minipage}
\caption{(a) Mean of filters from 12 layers from the setting MAE / FreqFiT-Bias / Caltech101, presented in Figure 3 in the main paper. (b) First 252 filters from the last layer of the same setting.
} \label{fig:filter_mae_bias}
\end{figure*}

\begin{figure*}[h]
\centering
\begin{minipage}[]{0.5\linewidth}
    \centering{MoCo / FreqFiT-Adapter / CIFAR100}
    \includegraphics[trim=55 15 40 15,clip,width=\textwidth]{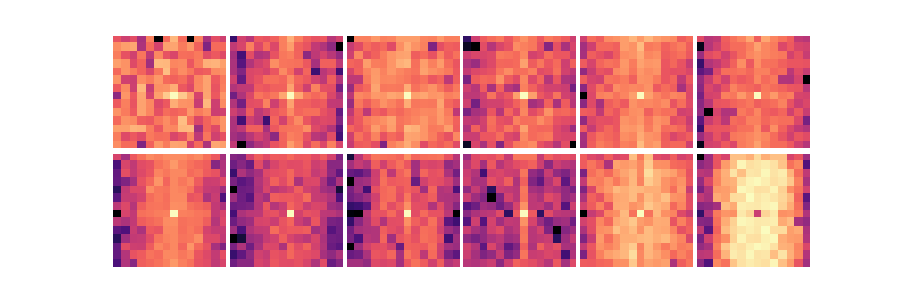} 
    \centering{(a) Mean of filters from 12 layers. \\Left to right, Upper row: layer 1-6. Lower row: layer 7-12}
\end{minipage}\\
\vspace{10pt}
\begin{minipage}[]{0.8\linewidth}
    \includegraphics[trim=80 15 80 15,clip,width=\textwidth]{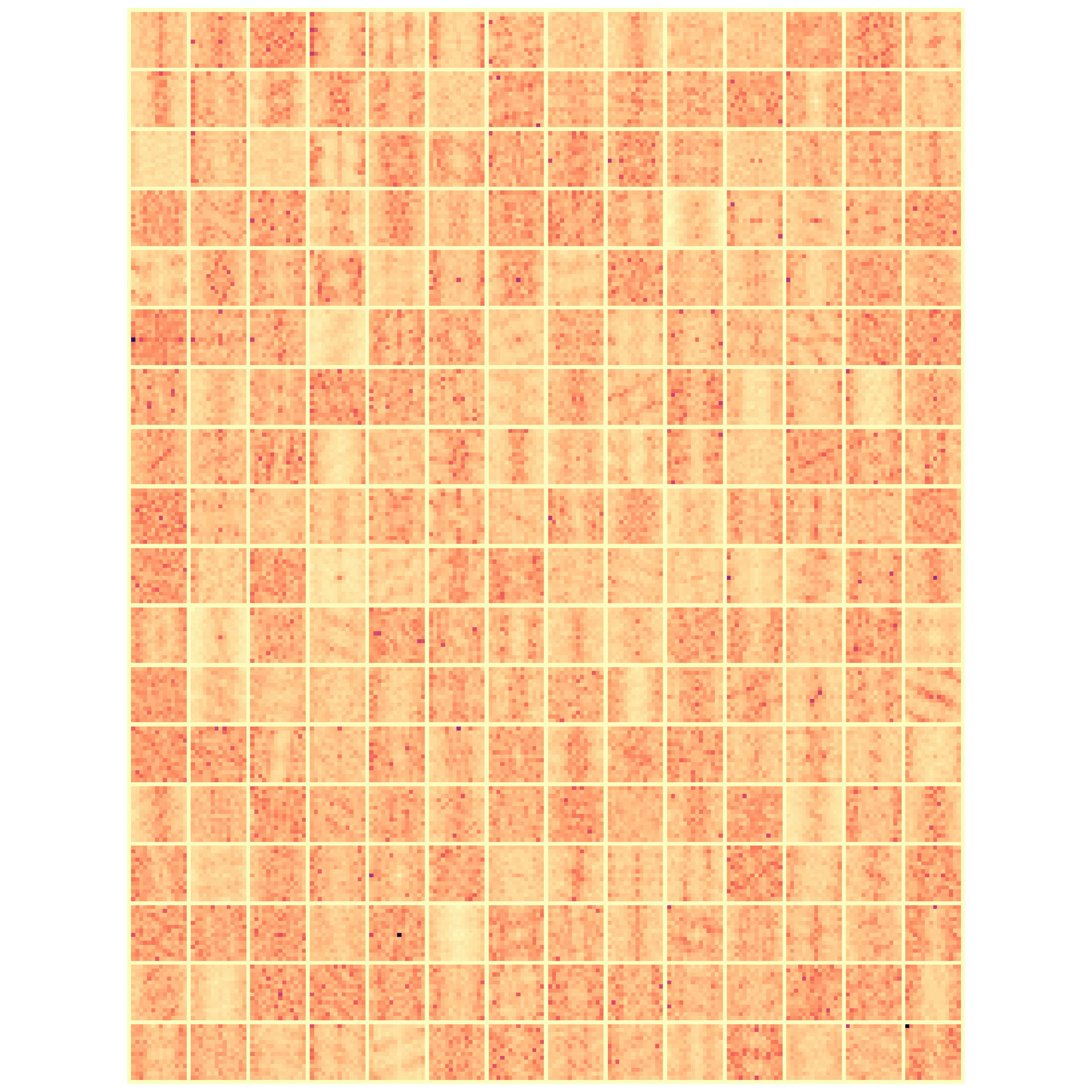} 
    \centering{(b) First 252 filters from the last layer}
\end{minipage}
\caption{(a) Mean of filters from 12 layers from the setting MoCo / FreqFiT-Adapter / CIFAR100, presented in Figure 3 in the main paper. (b) First 252 filters from the last layer of the same setting.
} \label{fig:filter_moco}
\end{figure*}

\begin{figure*}[h]
\centering
\begin{minipage}[]{0.5\linewidth}
    \centering{MoCo / FreqFiT-Bias / CIFAR100}
    \includegraphics[trim=55 15 40 15,clip,width=\textwidth]{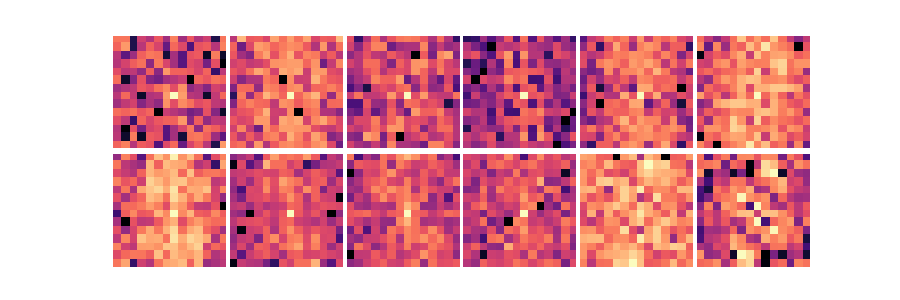} 
    \centering{(a) Mean of filters from 12 layers. \\Left to right, Upper row: layer 1-6. Lower row: layer 7-12}
\end{minipage}\\
\vspace{10pt}
\begin{minipage}[]{0.8\linewidth}
    \includegraphics[trim=80 15 80 15,clip,width=\textwidth]{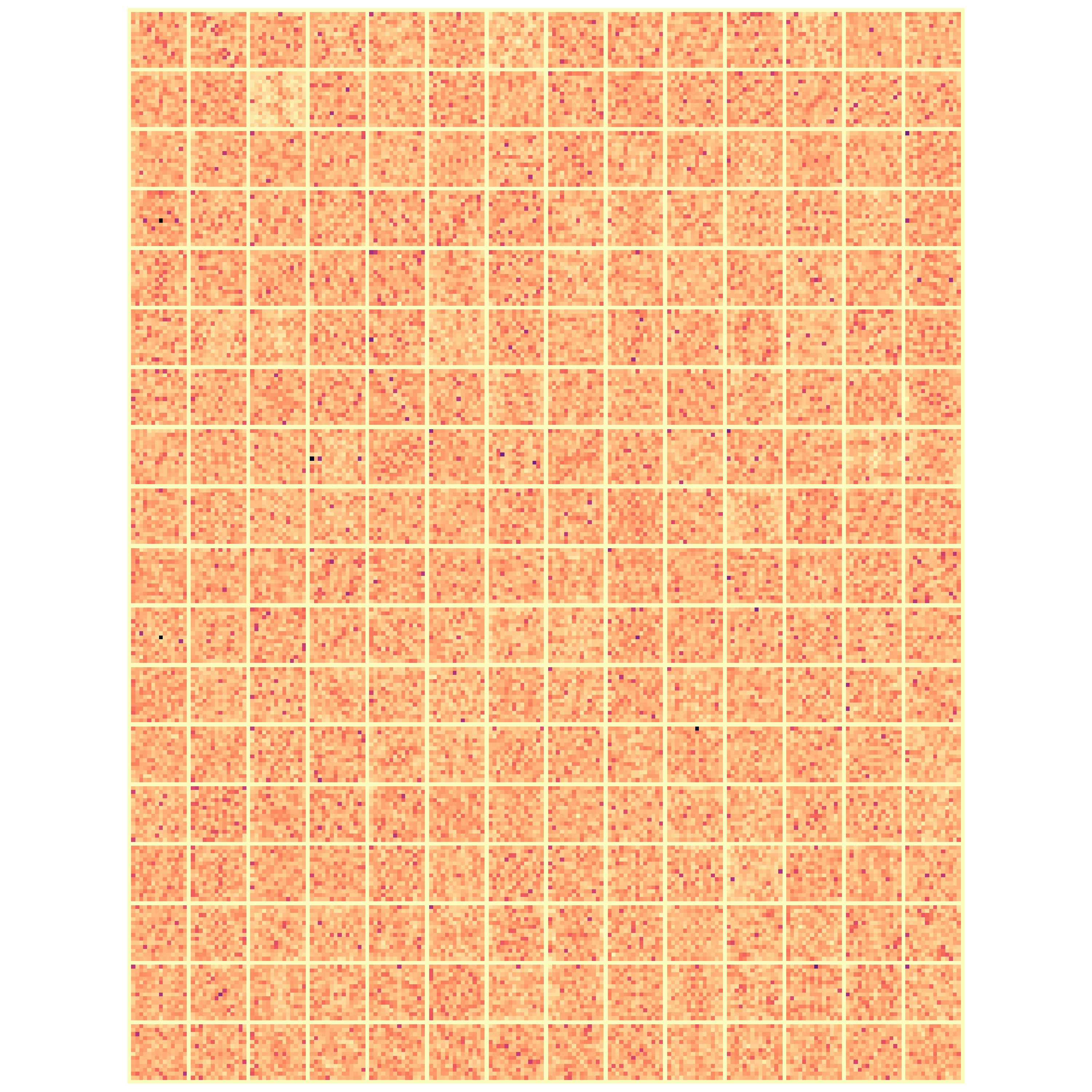} 
    \centering{(b) First 252 filters from the last layer.}
\end{minipage}
\caption{(a) Mean of filters from 12 layers from the setting MoCo / FreqFiT-Bias / CIFAR100, presented in Figure 3 in the main paper. (b) First 252 filters from the last layer of the same setting.
} \label{fig:filter_moco_bias}
\end{figure*}

\begin{figure*}[h]
\centering
\begin{minipage}[]{0.5\linewidth}
    \centering{CLIP / FreqFiT-VPT / Flower102}
    \includegraphics[trim=55 15 40 15,clip,width=\textwidth]{imgs/filters/clip_vpt_vtab_flower.eps} 
    \centering{(a) Mean of filters from 12 layers. \\Left to right, Upper row: layer 1-6. Lower row: layer 7-12}
\end{minipage}\\
\vspace{10pt}
\begin{minipage}[]{0.8\linewidth}
    \includegraphics[trim=80 15 80 15,clip,width=\textwidth]{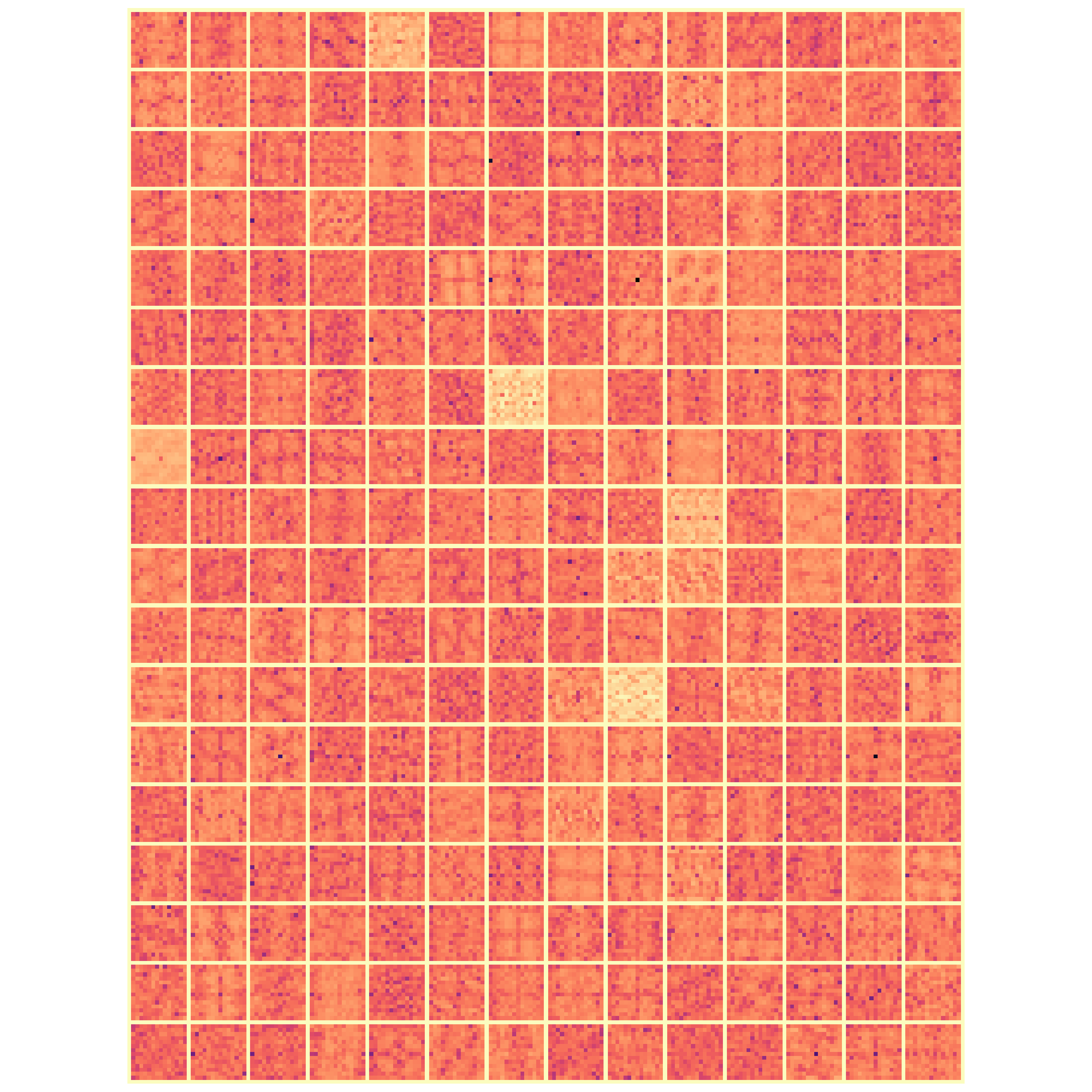} 
    \centering{(b) First 252 filters from the last layer.}
\end{minipage}
\caption{(a) Mean of filters from 12 layers from the setting CLIP / FreqFiT-VPT / Flower102, presented in Figure 3 in the main paper. (b) First 252 filters from the last layer of the same setting.
} \label{fig:filter_clip}
\end{figure*}

\begin{figure*}[h]
\centering
\begin{minipage}[]{0.5\linewidth}
    \centering{CLIP / FreqFiT-Bias / Flower102}
    \includegraphics[trim=55 15 40 15,clip,width=\textwidth]{imgs/filters/clip_bias_vtab_flower.eps} 
    \centering{(a) Mean of filters from 12 layers.\\Left to right, Upper row: layer 1-6. Lower row: layer 7-12}
\end{minipage}\\
\vspace{10pt}
\begin{minipage}[]{0.8\linewidth}
    \includegraphics[trim=80 15 80 15,clip,width=\textwidth]{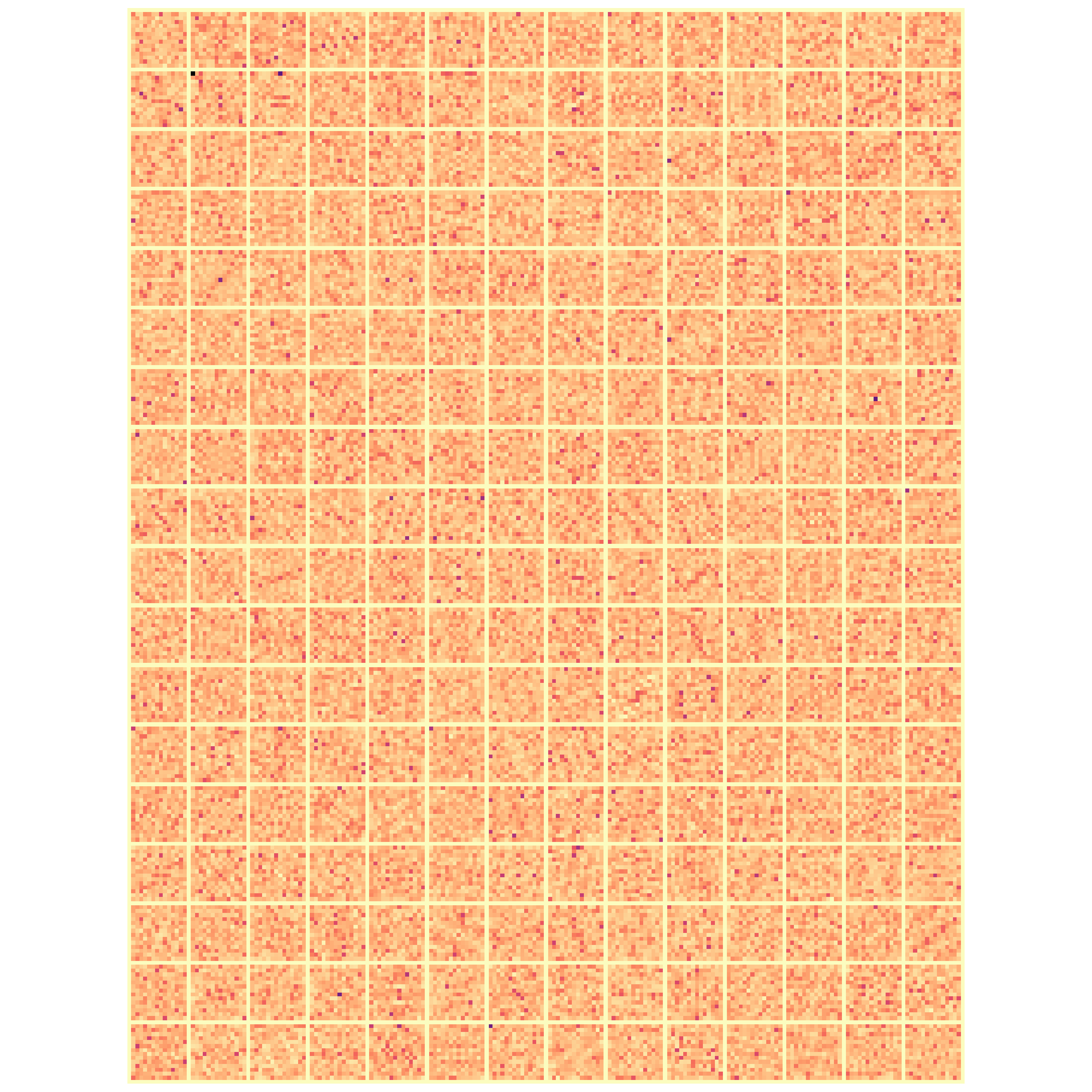} 
    \centering{(b) First 252 filters from the last layer.}
\end{minipage}
\caption{(a) Mean of filters from 12 layers from the setting CLIP / FreqFiT-Bias / Flower102, presented in Figure 3 in the main paper. (b) First 252 filters from the last layer of the same setting.
} \label{fig:filter_clip_bias}
\end{figure*}

\end{document}